\documentclass[runningheads,a4paper]{llncs}[2015/06/24]

\usepackage{cmap}
\usepackage{ownstyles}
\usepackage{array}
\usepackage{cellspace}
\usepackage{subcaption}
\usepackage{footmisc}

\iftrue 
  \usepackage[%
    rm={oldstyle=false,proportional=true},%
    sf={oldstyle=false,proportional=true},%
    tt={oldstyle=false,proportional=true,variable=true},%
    qt=false%
  ]{cfr-lm}
\else
  \usepackage{newtxtext}
  \usepackage{newtxmath}
  \usepackage[zerostyle=b,scaled=.9]{newtxtt}
\fi

\usepackage[math]{blindtext}

\usepackage[T1]{fontenc}
\usepackage[utf8]{inputenc} 

\usepackage{graphicx}
\usepackage{subcaption}
\usepackage{hyperref}       
\usepackage{amsmath}
\usepackage{amssymb}

\graphicspath{{./images/}}

\usepackage[ngerman,english]{babel}
\addto\extrasenglish{\languageshorthands{ngerman}\useshorthands{"}}

\usepackage[usenames,dvipsnames,svgnames,table]{xcolor}
\usepackage{hyperref}
\usepackage{backref}

\usepackage{cite}

\usepackage{upquote}

\usepackage{paralist}

\usepackage{csquotes}

\usepackage{mathtools}		  

\newif\ifcomments

\commentstrue

\ifcomments
\newcommand{\comments}[1]{#1}
\else
\newcommand{\comments}[1]{}
\fi

\newcommand{\layer}[1]{\ensuremath{\mathsf{#1}\xspace}}
\newcommand{\unit}[1]{\ensuremath{\mathsf{#1}\xspace}}

\newcommand{\thinplus}{\hspace*{-.3ex}+\hspace*{-.3ex}}

\newcommand{\thineq}{\hspace*{-.3ex}=\hspace*{-.3ex}}

\DeclareMathOperator*{\argmax}{arg\,max}
\newcommand{\x}{\mathbf{x}}

\newcommand{\h}{\mathbf{h}}

\newcommand{\R}{\mathbb{R}}

\RequirePackage{iftex}
\ifPDFTeX
  \RequirePackage[%
    final,%
    expansion=alltext,%
    protrusion=alltext-nott]{microtype}%
\else
  \RequirePackage[%
    final,%
    protrusion=alltext-nott]{microtype}%
\fi%

\usepackage{url}
\makeatletter
\g@addto@macro{\UrlBreaks}{\UrlOrds}
\makeatother

\usepackage{xcolor}

\usepackage{listings}
\renewcommand{\lstlistingname}{List.}

\let\counterwithout\relax

\usepackage{chngcntr}
\AtBeginDocument{\counterwithout{lstlisting}{section}}

\AtBeginEnvironment{listing}{\setcounter{listing}{\value{lstlisting}}}
\AtEndEnvironment{listing}{\stepcounter{lstlisting}}

\usepackage{pdfcomment}
\usepackage{hyperref}
\hypersetup{hidelinks,
  colorlinks=true,
  allcolors=black,
  pdfstartview=Fit,
  breaklinks=true}
\usepackage[all]{hypcap}

\usepackage[capitalise,nameinlink]{cleveref}
\crefname{section}{Sect.}{Sect.}
\Crefname{section}{Section}{Sections}
\crefname{listing}{\lstlistingname}{\lstlistingname}
\Crefname{listing}{Listing}{Listings}

\usepackage{ltxcmds}
\makeatletter
\newcommand{\IfPackageLoaded}[2]{\ltx@ifpackageloaded{#1}{#2}{}}
\makeatother

\IfPackageLoaded{minted}{
  \setminted{numbersep=5pt, xleftmargin=12pt}

  \usemintedstyle{friendly}

  \usepackage[labelfont=bf,font=small,skip=4pt]{caption}
  \SetupFloatingEnvironment{listing}{name=List.,within=none}
}{
}
\ifcsmacro{listing}{}{
  \newenvironment{listing}[1][htbp!]{\begin{figure}[#1]}{\end{figure}}
  \newcounter{listing}
}
\ifcsmacro{minted}{}{%
  
}

\usepackage{xspace}

\DeclareFontFamily{U}{MnSymbolC}{}
\DeclareSymbolFont{MnSyC}{U}{MnSymbolC}{m}{n}
\DeclareFontShape{U}{MnSymbolC}{m}{n}{
  <-6>    MnSymbolC5
  <6-7>   MnSymbolC6
  <7-8>   MnSymbolC7
  <8-9>   MnSymbolC8
  <9-10>  MnSymbolC9
  <10-12> MnSymbolC10
  <12->   MnSymbolC12%
}{}
\DeclareMathSymbol{\powerset}{\mathord}{MnSyC}{180}

\begin{document}

\title{Understanding Neural Networks \\via Feature Visualization: A survey}


 \author{%
     Anh Nguyen\inst{1} \and
     Jason Yosinski\inst{2} \and
     Jeff Clune\inst{2,3}\\
     \email{anhnguyen@auburn.edu}~~~\email{yosinski@uber.com}~~~\email{jeffclune@uwyo.edu}
 }
%
  \authorrunning{Anh Nguyen, Jason Yosinski, and Jeff Clune}
  \institute{
      Auburn University\\
      \and
      Uber AI Labs\\
      \and
      University of Wyoming
}

\maketitle

\begin{abstract}

A neuroscience method to understanding the brain is to find and study the \emph{preferred stimuli} that highly activate an individual cell or groups of cells.
Recent advances in machine learning enable a family of methods to synthesize preferred stimuli that cause a neuron in an artificial or biological brain to fire strongly.
Those methods are known as Activation Maximization (AM) \cite{erhan2009visualizing} or Feature Visualization via Optimization.
In this chapter, we (1) review existing AM techniques in the literature; (2) discuss a probabilistic interpretation for AM; and (3) review the applications of AM in debugging and explaining networks.

\end{abstract}


\section{Introduction}
\label{sec:intro}

\begin{figure}[h]
	\centering
	\includegraphics[width=0.6\columnwidth]{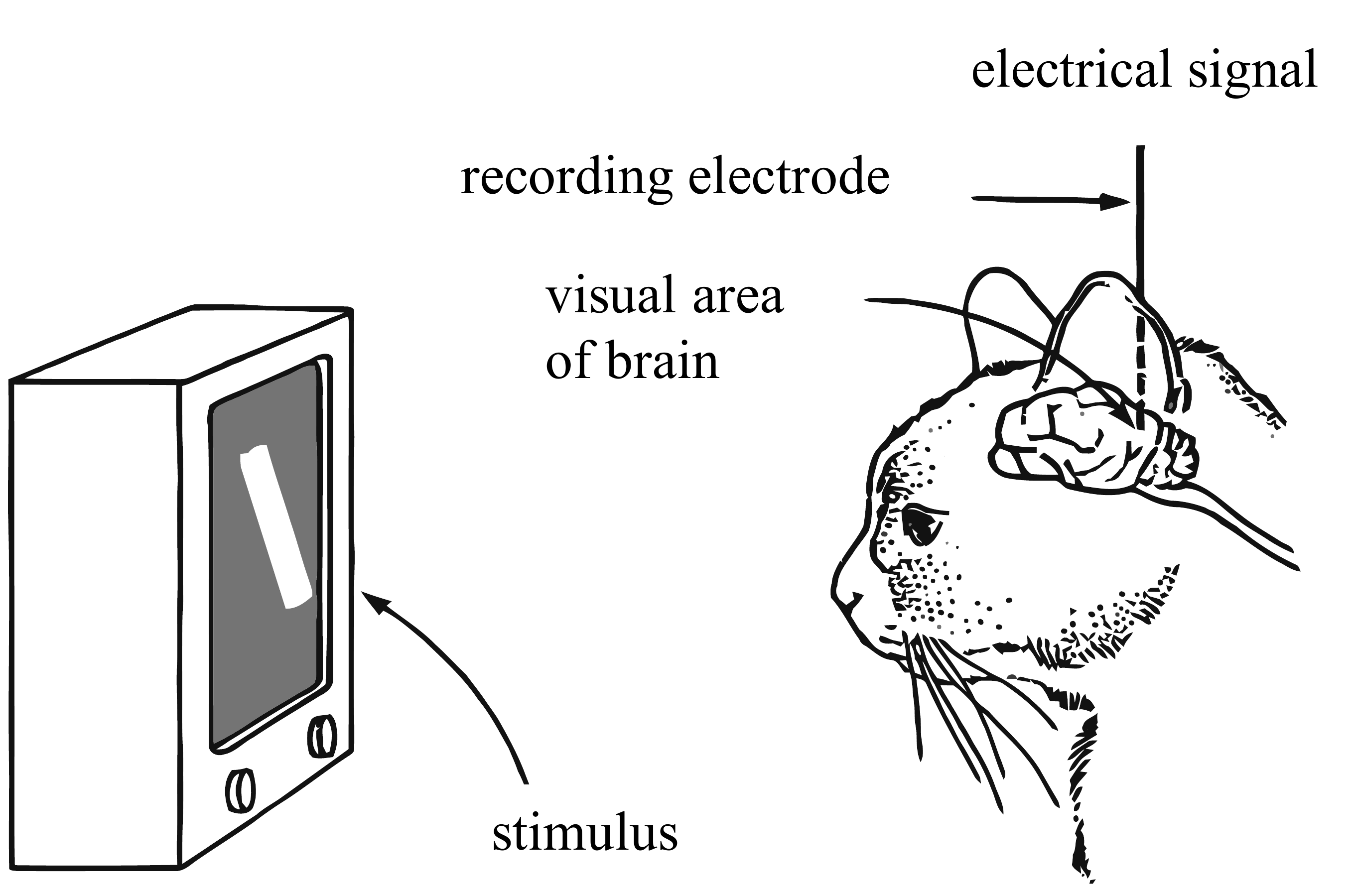}
	\caption{
		In the classic neuroscience experiment, Hubel and Wiesel discovered a cat's visual cortex neuron (right) that fires strongly and selectively for a light bar (left) when it is in certain positions and orientations \cite{hubel1959receptive}.
	}
	\vspace{-0.5cm}
	\label{fig:cat_experiment}
\end{figure}

Understanding the human brain has been a long-standing quest in human history.
One path to understanding the brain is to study what each neuron\footnote{In this chapter, ``neuron'', ``cell'', ``unit'', and ``feature'' are used interchangeably.} codes for \cite{kandel2000principles}, or what information its firing represents.
In the classic 1950's experiment, Hubel and Wiesel studied a cat's brain by showing the subject different images on a screen while recording the neural firings in the cat's primary visual cortex (Fig.~\ref{fig:cat_experiment}). 
Among a variety of test images, the researchers found \emph{oriented edges} to cause high responses in one specific cell \cite{hubel1959receptive}.
That cell is referred to as an \emph{edge detector} and such images are called its \emph{preferred stimuli}.
The same technique later enabled scientists to discover fundamental findings of how neurons along the visual pathway detect increasingly complex patterns: from circles, edges to faces and high-level concepts such as one's grandmother \cite{baer2007neuroscience} or specific celebrities like the actress Halle Berry \cite{quiroga2005invariant}.

Similarly, in machine learning (ML), visually inspecting the preferred stimuli of a unit can shed more light into what the neuron is doing \cite{zeiler2014visualizing,yosinski2015understanding}.
An intuitive approach is to find such preferred inputs from an existing, large image collection e.g. the training or test set \cite{zeiler2014visualizing}.
However, that method may have undesired properties. First, it requires testing each neuron on a large image set. 
Second, in such a dataset, many informative images that would activate the unit may not exist because the image space is vast and neural behaviors can be complex \cite{nguyen2015deep}.
Third, it is often ambiguous which visual features in an image are causing the neuron to fire
e.g. if a unit is activated by a picture of a bird on a tree branch, it is unclear if the unit ``cares about'' the bird or the branch (Fig.~\ref{fig:bird}b).
Fourth, it is not trivial how to extract a holistic description of what a neuron is for from the typically large set of stimuli preferred by a neuron.

A common practice is to study the top 9 highest activating images for a unit \cite{yosinski2015understanding,zeiler2014visualizing}; however, the top-9 set may reflect only one among many types of features that are preferred by a unit \cite{nguyen2016multifaceted}.

Instead of finding real images from an existing dataset, one can synthesize the visual stimuli from scratch \cite{olahfeature,erhan2009visualizing,nguyen2016synthesizing,nguyen2017ai,simonyan2013deep,wei2015understanding,nguyen2016multifaceted}.
The synthesis approach offers multiple advantages: (1) given a strong image prior, one may synthesize (i.e. reconstruct) stimuli without the need to access the target model's training set, which may not be available in practice (see Sec.~\ref{sec:results}); (2) more control over the types and contents of images to synthesize, which helps shed light on more controlled research experiments.


\noindent\textbf{Activation Maximization}
Let $\theta$ be the parameters of a classifier that maps an image $\x \in \R^{H\times W\times C}$ (that has $C$ color channels, each of which is $W$ pixels wide and $H$ pixels high) onto a probability distribution over the output classes.
Finding an image $\x$ that maximizes the activation $a^{l}_i(\theta, \x)$ of a neuron indexed $i$ in a given layer $l$ of the classifier network can be formulated as an optimization problem:

\begin{equation}
\label{eq:activation_maximization}
\x^* = \argmax_{\x}(a^{l}_i(\theta, \x))
\end{equation}

\noindent This problem was introduced as \emph{activation maximization}\footnote{Also sometimes referred to as \emph{feature visualization} \cite{olahfeature,nguyen2016multifaceted,yosinski2015understanding}.
	In this chapter, the phrase ``visualize a unit'' means ``synthesize preferred images for a single neuron''.
} (AM) by Erhan, Bengio and others \cite{erhan2009visualizing}.
Here, $a^{l}_i(.)$ returns the activation value of a \emph{single} unit as in many previous works \cite{nguyen2015deep,nguyen2016synthesizing,nguyen2016multifaceted}; however, it can be extended to return any neural response $a(.)$ that we wish to study e.g. activating a group of neurons \cite{mordvintsev2015inceptionism,olah2018building,nguyen2017plug}.
The remarkable DeepDream visualizations \cite{mordvintsev2015inceptionism} were created by running AM to activate all the units across a given layer simultaneously.
In this chapter, we will write $a(.)$ instead of $a^l_i(.)$ when the exact indices $l,i$ can be omitted for generality.


AM is a non-convex optimization problem for which one can attempt to find a local minimum via gradient-based \cite{szegedy2013intriguing-properties-of-neural} or non-gradient methods \cite{nguyen2016understanding}.
In \emph{post-hoc} interpretability \cite{montavon2017methods}, we often assume access to the parameters and architecture of the network being studied.
In this case, a simple approach is to perform gradient ascent \cite{yosinski2015understanding,erhan2009visualizing,nguyen2016synthesizing,nguyen2015innovation} with an update rule such as:

\begin{equation}
\x_{t+1} = \x_{t} + 
\epsilon_1 \frac{\partial a(\theta, \x_{t})}{\partial \x_{t}}
\label{eqn:am_no_priors}
\end{equation}

\noindent That is, starting from a random initialization $\x_0$ (here, a random image), we iteratively take steps in the input space following the gradient of $a(\theta, \x)$ to find an input $\x$ that highly activates a given unit.
$\epsilon_1$ is the step size and is chosen empirically.

Note that this gradient ascent process is similar to the gradient descent process used to train neural networks via backpropagation \cite{rumelhart1986learning}, except that here we are optimizing the network input instead of the network parameters $\theta$, which are frozen.\footnote{Therefore, hereafter, we will write $a(x)$ instead of $a(\theta, x)$, omitting  $\theta$, for simplicity.}
We may stop the optimization when 
the neural activation has reached a desired threshold or a certain number of steps has passed.

\begin{figure}[!h]
	\centering
	\includegraphics[width=0.5\columnwidth]{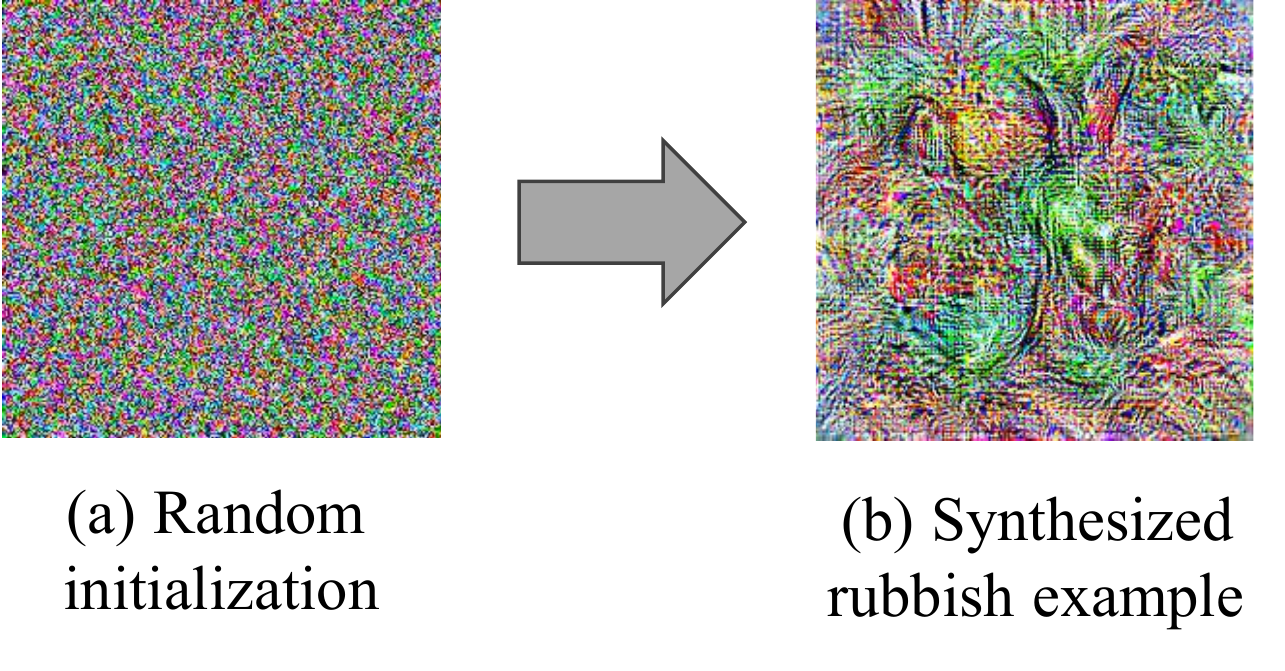}
	\vspace*{-0.5em}
	\caption{
		Example of activation maximization without image priors.
		Starting from a random image (a), we iteratively take steps following the gradient to maximize the activation of a given unit, here the ``bell pepper'' output in CaffeNet \cite{krizhevsky2012imagenet}.
		Despite highly activating the unit and being classified as ``bell pepper'', the image (b) has high frequencies and is not human-recognizable.
	}
	\label{fig:rubbish}
\end{figure}

In practice, synthesizing an image from scratch to maximize the activation alone (i.e. an unconstrained optimization problem) often yields uninterpretable images \cite{nguyen2015deep}.
In a high-dimensional image space, we often find \emph{rubbish} examples (also known as \emph{fooling} examples \cite{nguyen2015deep}) e.g. patterns of high-frequency noise that look like nothing but that highly activate a given unit (Fig.~\ref{fig:rubbish}).

In a related way, if starting AM optimization from a real image (instead of a random one), we may easily encounter \emph{adversarial} examples \cite{szegedy2013intriguing-properties-of-neural} e.g. an image that is slightly different from the starting image (e.g. of a school bus), but that a network would give an entirely different label e.g. ``ostrich'' \cite{szegedy2013intriguing-properties-of-neural}.
Those early AM visualizations \cite{szegedy2013intriguing-properties-of-neural,nguyen2015deep} revealed huge security and reliability concerns with machine learning applications and informed a plethora of follow-up adversarial attack and defense research \cite{akhtar2018threat,kabilan2018vectordefense}.
\\

\noindent\textbf{Networks that we visualize} Unless otherwise noted, throughout the chapter, we demonstrate AM on CaffeNet, a specific pre-trained model of the well-known AlexNet convnets \cite{krizhevsky2012imagenet} to perform single-label image classification on the ILSVRC 2012 ImageNet dataset
\cite{deng2009imagenet, russakovsky2014imagenet}.

\section{Activation Maximization via Hand-designed Priors}
\label{sec:am_priors}

Examples like those in Fig.~\ref{fig:rubbish}b are not human-recognizable.
While the fact that the network responds strongly to such images is intriguing and has strong implications for security, if we cannot interpret the images, it limits our ability to understand what the unit's purpose is.
Therefore, we want to constrain the search to be within a distribution of images that we can interpret e.g. photo-realistic images or images that look like those in the training set.
That can be accomplished by incorporating \emph{natural image priors} into the objective function, which was found to substantially improve the recognizability of AM images \cite{yosinski2015understanding,mahendran2015visualizing,nguyen2016multifaceted,nguyen2016synthesizing,olahfeature}.
For example, an image prior may encourage smoothness \cite{mahendran2015visualizing} or penalize pixels of extreme intensity \cite{simonyan2013deep}.
Such constraints are often incorporated into the AM formulation as a \emph{regularization} term $R(\x)$:

\begin{equation}
\label{eq:am_prior}
\x^* = \argmax_{\x}(a(\x) - R(\x))
\end{equation}

For example, to encourage the smoothness in AM images, $R: \R^{H\times W\times C} \to \R$ may compute the total variation (TV) across an image \cite{mahendran2015visualizing}.
That is, in each update, we follow the gradients to (1) maximize the neural activation; and (2) minimize the total variation loss:

\begin{equation}
\x_{t+1} = \x_{t} + 
\epsilon_1 \frac{\partial a(\x_{t})}{\partial \x_{t}} -
\epsilon_2 \frac{\partial R(\x_{t})}{\partial \x_{t}}
\end{equation}

\noindent However, in practice, we do not always compute the analytical gradient $\partial R(\x_{t})/\partial \x_t$.
Instead, we may define a regularization operator $r:\R^{H\times W\times C} \to \R^{H\times W\times C}$ (e.g. a Gaussian blur kernel), and map $\x$ to a more regularized (e.g. slightly blurrier as in \cite{yosinski2015understanding}) version of itself in every step. 
In this case, the update step becomes:

\begin{equation}
\x_{t+1} = r(\x_{t}) + 
\epsilon_1 \frac{\partial a(\x_{t})}{\partial \x_{t}}
\label{eqn:r}
\end{equation}

Note that this update form in Eq.~\ref{eqn:r} is strictly more expressive \cite{yosinski2015understanding}, and allows the use of non-differentiable regularizers $r(.)$.\\


\begin{figure}[t!h]
	\centering
	\includegraphics[width=1.0\columnwidth]{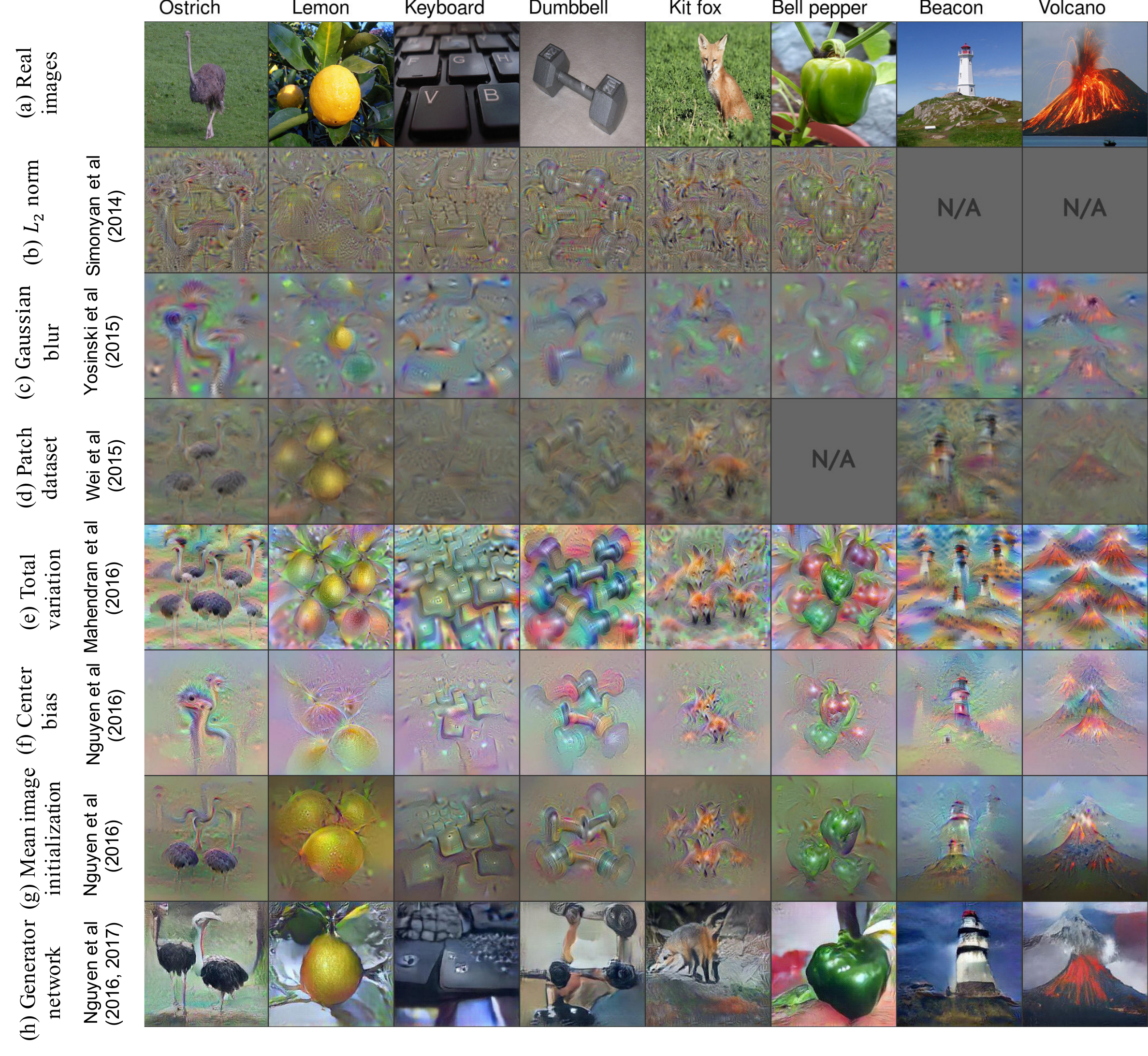}
	\caption{
		Activation maximization results of seven methods in the literature (b--h), each employing a different image prior (e.g. $L_2$ norm, Gaussian blur, etc.).
		Images are synthesized to maximize the output neurons (each corresponding to a class) of the CaffeNet image classifier \cite{krizhevsky2012imagenet} trained on ImageNet.
		The categories were not cherry-picked, but instead were selected based on the images available in previous papers~\cite{simonyan2013deep,yosinski2015understanding,wei2015understanding,nguyen2016multifaceted,mahendran2015visualizing}. Overall, while it is a subjective judgement, Activation Maximization via Deep Generator Networks method (h) \cite{nguyen2016synthesizing} produces images with more natural colors and realistic global structures.
		Image modified from \cite{nguyen2016synthesizing}.
	}
	\label{fig:prev_work}
\end{figure}

\noindent\textbf{Local statistics} 
AM images without priors often appear to have high-frequency patterns and unnatural colors (Fig.~\ref{fig:rubbish}b).
Many regularizers have been designed in the literature to ameliorate these problems including:

\begin{itemize}
	\item Penalize extreme-intensity pixels via $\alpha$-norm~\cite{simonyan2013deep,yosinski2015understanding,wei2015understanding} (Fig.~\ref{fig:prev_work}b).
	\item Penalize high-frequency noise (i.e. smoothing) via total variation \cite{mahendran2015visualizing,nguyen2016multifaceted} (Fig.~\ref{fig:prev_work}e), Gaussian blurring \cite{yosinski2015understanding,oygard2015visualizing} (Fig.~\ref{fig:prev_work}c) or a bilateral filter \cite{tyka2016bilateral}.
	\item Randomly jitter, rotate, or scale the image before each update step to synthesize stimuli that are robust to transformations, which has been shown to make images clearer and more interpretable \cite{olahfeature,mordvintsev2015inceptionism}.
	\item Penalize the high frequencies in the \emph{gradient} image $\frac{\partial a(\x_{t})}{\partial \x_{t}}$ (instead of the visualization $\x_{t}$) via Gaussian blurring \cite{oygard2015visualizing,olahfeature}.
	\item Encourage patch-level color statistics to be more realistic by (1) matching those of real images from a dataset \cite{wei2015understanding} (Fig.~\ref{fig:prev_work}d) or (2) learning a Gaussian mixture model of real patches \cite{mordvintsev2015inceptionism}.
\end{itemize}

\noindent While substantially improving the interpretability of images (compared to high-frequency rubbish examples), these methods only effectively attempt to match the \emph{local} statistics of natural images.\\

\noindent\textbf{Global structures}
Many AM images still lack \emph{global} coherence; for example, an image synthesized to
highly activate the ``bell pepper'' output neuron (Fig.~\ref{fig:prev_work}b--e) may exhibit multiple bell-pepper segments scattered around the same image rather than a single bell pepper.
Such stimuli suggest that the network has learned some \emph{local} discriminative features e.g. the shiny, green skin of bell peppers, which are useful for the classification task.
However, it raises an interesting question: Did the network ever learn the global structures (e.g. the whole pepper) or only the local discriminative parts?
The high-frequency patterns as in Fig.~\ref{fig:prev_work}b--e might also be a consequence of optimization in the image space.
That is, when making pixel-wise changes, it is non-trivial to ensure global coherence across the entire image.
Instead, it is easy to increase neural activations by simply creating more local discriminative features in the stimulus.

Previous attempts to improve the global coherence include:

\begin{itemize}
	\item Gradually paint the image by scaling it and alternatively following the gradients from multiple output layers of the network \cite{oygard2015visualizing}.
	\item Bias the image changes to be near the image center \cite{nguyen2016multifaceted} (Fig.~\ref{fig:prev_work}g).
	\item Initialize optimization from an average image (computed from real training set images) instead of a random one \cite{nguyen2016multifaceted} (Fig.~\ref{fig:prev_work}h).
\end{itemize}

While these methods somewhat improved the global coherence of images (Fig.~\ref{fig:prev_work}g--h), they rely on a variety of heuristics and introduce extra hyperparameters \cite{oygard2015visualizing,nguyen2016multifaceted}.
In addition, there is still a large realism gap between the real images and these visualizations (Fig.~\ref{fig:prev_work}a vs. h).\\

\noindent\textbf{Diversity}
A neuron can be multifaceted in that it responds strongly to multiple distinct types of stimuli, i.e. \emph{facets} \cite{nguyen2016multifaceted}.
That is, higher-level features are more invariant to changes in the input \cite{zeiler2014visualizing,le2013building}.
For example, a face-detecting unit in CaffeNet \cite{krizhevsky2012imagenet} was found to respond to both human and lion faces \cite{yosinski2015understanding}. 
Therefore, we wish to uncover different facets via AM in order to have a fuller understanding of a unit.

However, AM optimization starting from different random images often converge to similar results \cite{erhan2009visualizing,nguyen2016multifaceted}---a phenomenon also observed when training neural networks with different initializations \cite{li2015convergent}.
Researchers have proposed different techniques to improve image diversity such as:

\begin{itemize}
	\item Drop out certain neural paths in the network when performing backpropagation to produce different facets \cite{wei2015understanding}.
	\item Cluster the training set images into groups, and initialize from an average image computed from each group's images \cite{nguyen2016multifaceted}.
	\item Maximize the distance (e.g. cosine similarity in the pixel space) between a reference image and the one being synthesized \cite{olahfeature}.
	\item Activate two neurons at the same time e.g. activating (bird + apron) and (bird + candles) units would produce two distinct images of \emph{birds} that activate the same \emph{bird} unit \cite{nguyen2016synthesizing} (Fig.~\ref{fig:am_math}).
	\item Add noise to the image in every update to increase image diversity \cite{nguyen2017plug}.
\end{itemize}

While obtaining limited success, these methods also introduce extra hyperparameters and require further investigation. For example, if we enforce two stimuli to be different, exactly how far should they be and in which similarity metric should the difference be measured?

\section{Activation Maximization via Deep Generator Networks}
\label{sec:dgnam}

\begin{figure}[t!h]
	\centering
	\includegraphics[width=0.9\columnwidth]{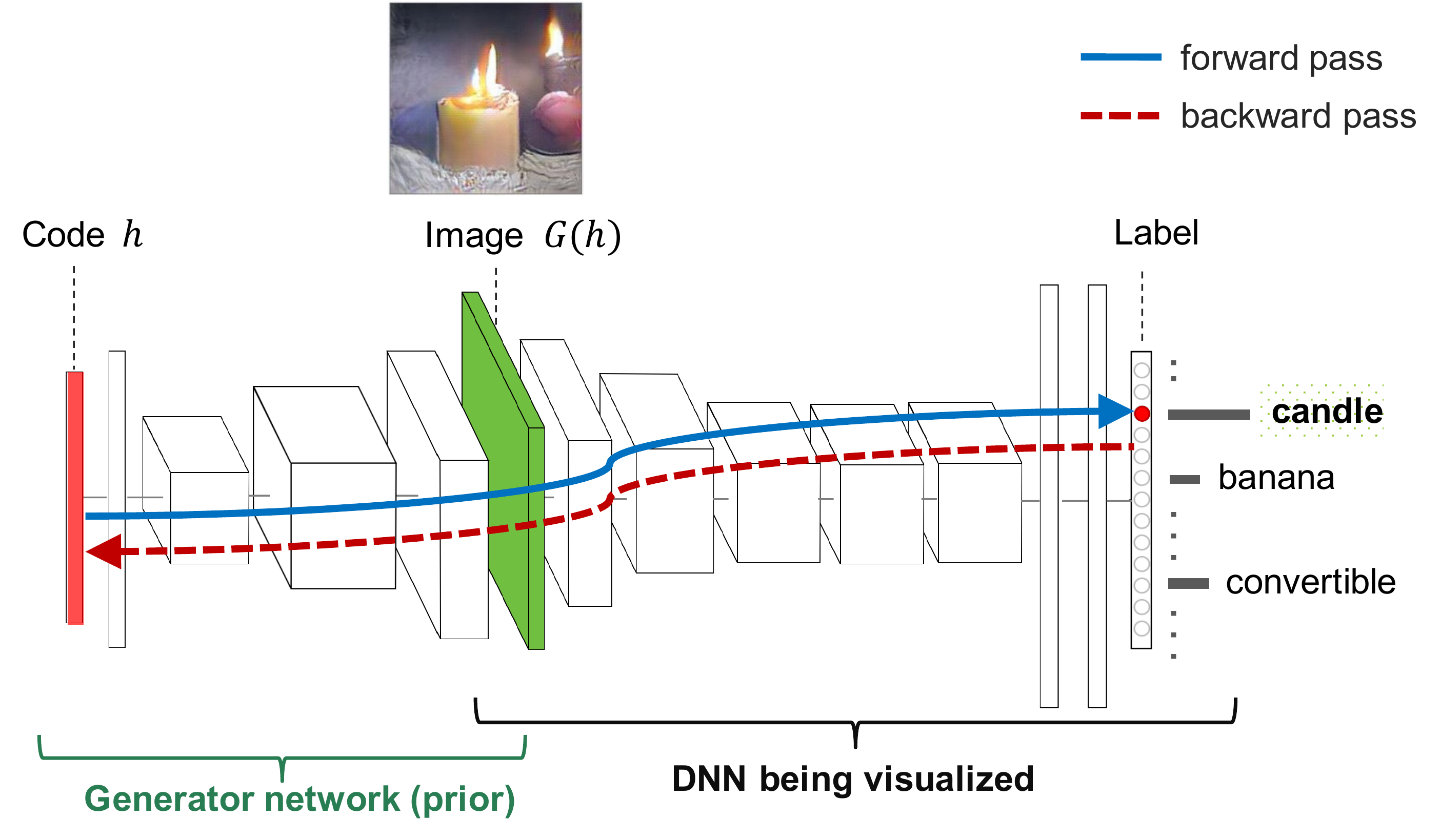}
	\caption{
		We search for an input code (red bar) of a deep generator network (left) that produces an image (middle) that strongly activates a target neuron (e.g. the ``candle'' output unit) in a given pre-trained network (right). 
		The iterative optimization procedure involves multiple forward and backward passes through both the generator and the target network being visualized.
	}
	\label{fig:main_concept}
\end{figure}

Much previous AM research were optimizing the preferred stimuli directly in the high-dimensional image space where pixel-wise changes are often slow and uncorrelated, yielding high-frequency visualizations (Fig.~\ref{fig:prev_work}b--e).
Instead, Nguyen et al. \cite{nguyen2016synthesizing} propose to optimize in the low-dimensional latent space of a deep generator network, which they call Deep Generator Network Activation Maximization (DGN-AM).
They train an image \emph{generator} network to take in a highly compressed code 
and output a synthetic image that looks as close to real images from the ImageNet dataset~\cite{russakovsky2014imagenet} as possible.
To produce an AM image for a given neuron, the authors optimize in the input latent space of the generator so that it outputs an image that activates the unit of interest (Fig.~\ref{fig:main_concept}). 
Intuitively, DGN-AM restricts the search to only the set of images that can be drawn by the prior and encourages the image updates to be more coherent and correlated compared to pixel-wise changes (where each pixel is modified independently).\\

\noindent\textbf{Generator networks}
We denote the sub-network of CaffeNet \cite{krizhevsky2012imagenet} that maps images onto 4096-D \layer{fc6} features as an encoder $E:\R^{H\times W\times C} \to \R^{4096}$.
We train a generator network $G: \R^{4096} \to \R^{H\times W\times C}$ to invert $E$ i.e. $G(E(\x)) \approx \x$.
In addition to the reconstruction losses, the generator was trained using the Generative Adversarial Network (GAN) loss \cite{goodfellow2014generative} to improve the image realism.
More training details are in \cite{nguyen2016synthesizing,dosovitskiy2016generating}.
Intuitively, $G$ can be viewed as an artificial \emph{general} ``painter'' that is capable of painting a variety of different types of images, given an arbitrary input description (i.e. a latent code or a condition vector).
The idea is that $G$ would be able to faithfully portray what a target network has learned, which may be recognizable or unrecognizable patterns to humans.

\noindent\textbf{Optimizing in the latent space} Intuitively, we search in the input code space of the generator $G$ to find a code $\h \in \R^{4096}$ such that the image $G(\h)$ maximizes the neural activation $a(G(\h))$ (see Fig.~\ref{fig:main_concept}). 
The AM problem in Eq.~\ref{eq:am_prior} now becomes:

\begin{equation}
\label{eq:am_prior_latent}
\h^* = \argmax_{\h}(a(G(\h)) - R(\h))
\end{equation}

\noindent That is, we take steps in the latent space following the below update rule:

\begin{equation}
\h_{t+1} = \h_{t} + 
\epsilon_1 \frac{\partial a(G(\h_t))}{\partial \h_{t}} -
\epsilon_2 \frac{\partial R(\h_{t})}{\partial \h_{t}}
\end{equation}

Note that, here, the regularization term $R(.)$ is on the latent code $\h$ instead of the image $\x$.
Nguyen et al. \cite{nguyen2016synthesizing} implemented a small amount of $L_2$ regularization and also clipped the code. 
These hand-designed regularizers can be replaced by a strong, learned prior for the code \cite{nguyen2017plug}.

\begin{figure}[t!h]
	\centering
	\vspace*{0.55cm}
	\includegraphics[width=1.0\columnwidth]{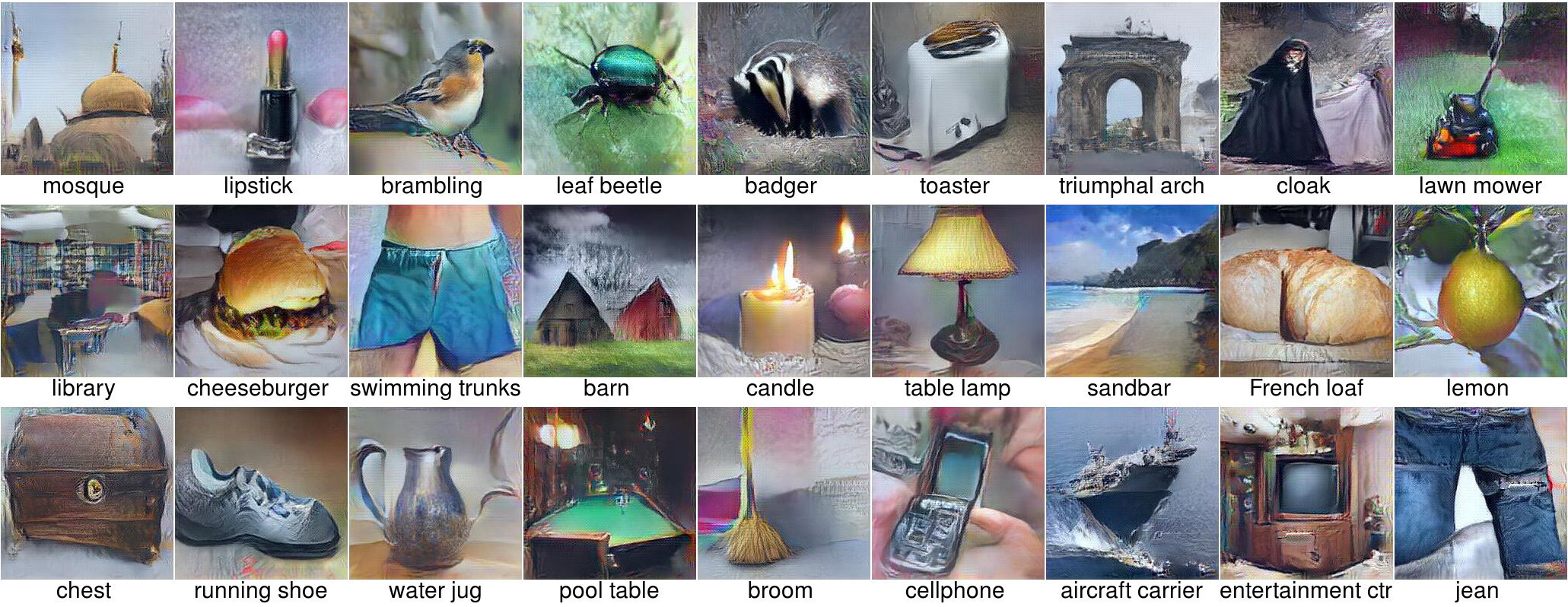}
	\caption{
		Images synthesized from scratch via DGN-AM method \cite{nguyen2016synthesizing} to highly activate output neurons in the CaffeNet deep neural network \cite{jia2014caffe}, which has learned to classify 1000 categories of ImageNet images. Image from \cite{nguyen2016synthesizing}.
	}
	\label{fig:teaser}
\end{figure}

\begin{figure}
	\centering
	\includegraphics[width=1.0\columnwidth]{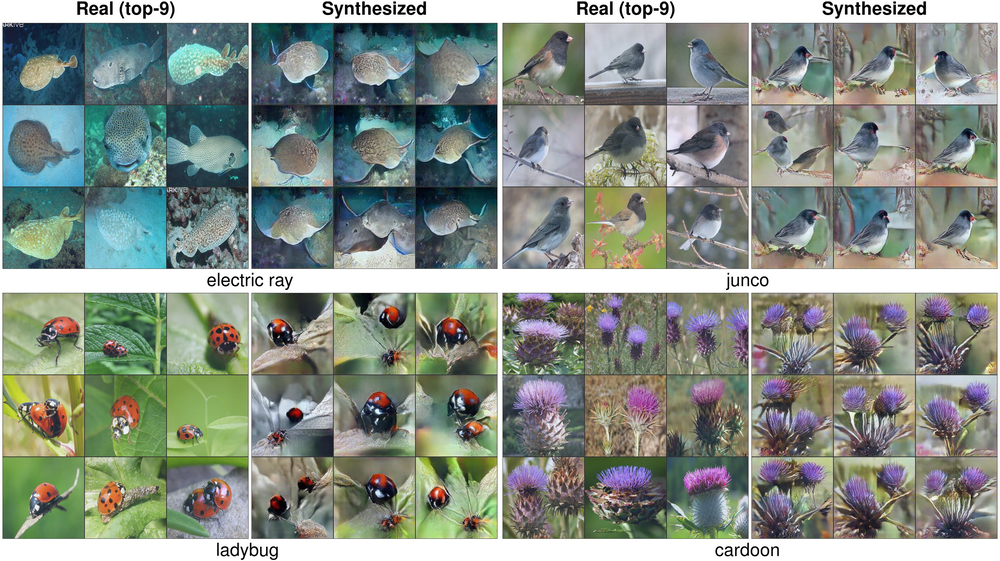}
	\caption{
		Side-by-side comparison between real and synthetic stimuli synthesized via DGN-AM \cite{nguyen2016synthesizing}. For each unit, we show the top 9 validation set images that highest activate a given neuron (left) and 9 synthetic images (right). Note that these synthetic images are of size $227\times227$ i.e. the input size of CaffeNet \cite{krizhevsky2012imagenet}.
		Image from \cite{nguyen2016synthesizing}.
		\vspace*{-0.5cm}
	}
	\label{fig:real_vs_synthesized}
\end{figure}

Optimizing in the latent space of a deep generator network showed a great improvement in image quality compared to previous methods that optimize in the pixel space (Fig.~\ref{fig:teaser}; and Fig.~\ref{fig:prev_work}b--h vs. Fig.~\ref{fig:prev_work}i).
However, images synthesized by DGN-AM have limited diversity---they are qualitatively similar to the real top-9 validation images that highest activate a given unit (Fig.~\ref{fig:real_vs_synthesized}).

To improve the image diversity, Nguyen et al. \cite{nguyen2017plug} harnessed a learned realism prior for $\h$ via a denoising autoencoder (DAE), and added a small amount of Gaussian noise in every update step to improve image diversity \cite{nguyen2017plug}.
In addition to an improvement in image diversity, this AM procedure also has a theoretical probabilistic justification, which is discussed in Section~\ref{sec:probabilistic}.

\section{Probabilistic interpretation for Activation Maximization}
\label{sec:probabilistic}

In this section, we first make a note about the AM objective, and discuss a probabilistically interpretable formulation for AM, which is first proposed in Plug and Play Generative Networks (PPGNs) \cite{nguyen2017plug}, and then interpret other AM methods under this framework.
Intuitively, the AM process can be viewed as sampling from a generative model, which is composed of (1) an image prior and (2) a recognition network that we want to visualize.

\subsection{Synthesizing selective stimuli}

We start with a discussion on AM objectives.
In the original AM formulation (Eq.~\ref{eq:activation_maximization}), we only explicitly maximize the activation $a^{l}_{i}$ of a unit indexed $i$ in layer $l$; however, in practice, this objective may surprisingly also increase the activations $a^{l}_{j\ne i}$ of some other units $j$ in the same layer and even higher than $a^{l}_i$ \cite{nguyen2016synthesizing}. 
For example, maximizing the output activation for the ``hartebeest'' class is likely to yield an image that also strongly activates the ``impala'' unit because these two animals are visually similar \cite{nguyen2016synthesizing}.
As the result, there is no guarantee that the target unit will be the highest activated across a layer.
In that case, the resultant visualization may not portray what is unique about the target unit $(l,i)$.

Instead, we are interested in \emph{selective} stimuli that highly activate only $a^l_i$, but not $a^l_{j\ne i}$.
That is, we wish to maximize $a^l_i$ such that it is the highest single activation across the same layer $l$.
To enforce that selectivity, we can either maximize the softmax or log of softmax of the raw activations across a layer \cite{simonyan2013deep,nguyen2017plug} where the softmax transformation for unit $i$ across layer $l$ is given as $s^l_i = \exp(a^l_i) / \sum_j \exp(a^l_j)$.
Such selective stimuli (1) are more interpretable and preferred in neuroscience \cite{baer2007neuroscience} because they contain only visual features exclusively for one unit of interest but not others; (2) naturally fit in our probabilistic interpretation discussed below.



\subsection{Probabilistic framework}
\label{sec:framework}

Let us assume a joint probability distribution $p(\x,y)$ where $\x$ denotes images, and $y$ is a categorical variable for a given neuron indexed $i$ in layer $l$.
This model can be decomposed into an image density model and an image classifier model:

\beq
p(\x, y) = p(\x) p(y | \x)
\eeq


Note that, when $l$ is the output layer of an ImageNet 1000-way classifier \cite{krizhevsky2012imagenet}, $y$ also represents the image category (e.g. ``volcano''), and $p(y | \x)$ is the classification probability distribution (often modeled via softmax).




We can construct a Metropolis-adjusted Langevin
\cite{roberts1998optimal} (MALA) sampler for our $p(\x,y)$ model \cite{nguyen2017plug}.
This variant of MALA \cite{nguyen2017plug} does not have the accept/reject step, and uses the following transition operator:\footnote{We abuse notation slightly in the interest of space and denote as $N(0, \epsilon_3^2)$ a sample from that distribution. The first step size is given as $\epsilon_{12}$ in anticipation of later splitting into separate $\epsilon_1$ and $\epsilon_2$ terms.}

\beq
\x_{t+1} = \x_t + \epsilon_{12}\nabla\log p(\x_t, y) + N(0, \epsilon_3^2)
\eqnlabel{mala-approx}
\eeq

\noindent Since $y$ is a categorical variable, and chosen to be a fixed neuron $y_c$ outside the sampler, the above update rule can be re-written as:

\beq
\x_{t+1} = \x_t \thinplus \epsilon_{12}\nabla\log p(y \thineq y_c | \x_t) \thinplus \epsilon_{12}\nabla\log p(\x_t ) \thinplus N(0, \epsilon_3^2)
\eeq

\noindent Decoupling $\epsilon_{12}$ into explicit $\epsilon_1$ and $\epsilon_2$ multipliers, and expanding the $\nabla$ into explicit partial derivatives, we arrive at the following update rule:

\beq
\x_{t+1} = \x_t + \epsilon_1\frac{\partial \log p(y = y_c | \x_t)}{\partial \x_t} + \epsilon_2\frac{\partial\log p(\x_t)}{\partial{\x_t}} + N(0, \epsilon_3^2)
\eqnlabel{update_rule}
\eeq

\begin{figure}[t!h]
	\centering
	\vspace*{-0.5cm}
	\includegraphics[width=0.84\columnwidth]{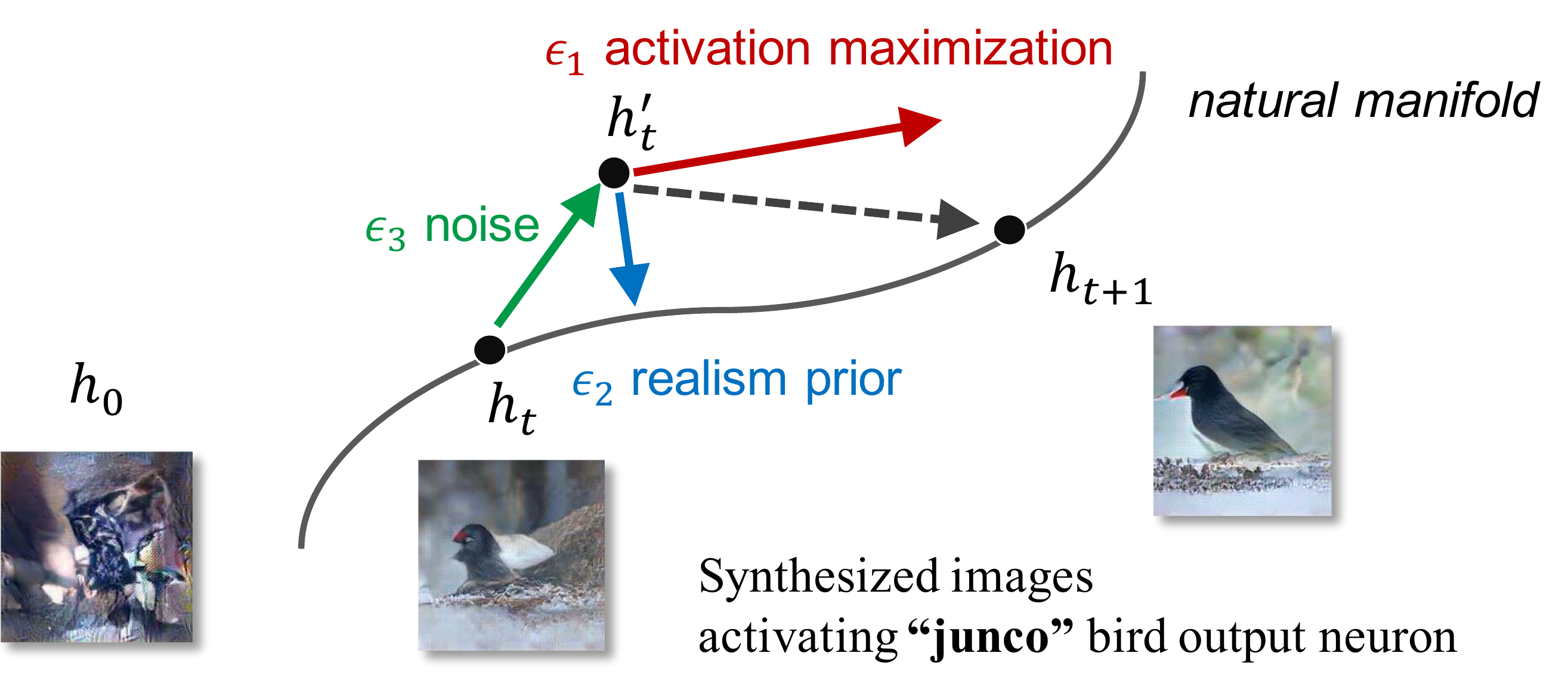}
	\caption{
		AM can be considered as a sampler, traversing in the natural image manifold.
		We start from a random initialization $h_0$.
		In every step $t$, we first add a small amount of noise (green arrow), which pushes the sample off the natural-image manifold ($h^{\prime}_t$).
		The gradients toward maximizing activation (red arrow) and more realistic images (blue arrow) pull the noisy $h^{\prime}_t$ back to the manifold at a new sample $h_{t+1}$.
	}
	\vspace*{-0.3cm}
	\label{fig:ppgn_update}
\end{figure}

An intuitive interpretation of the roles of these three terms is illustrated in Fig.~\ref{fig:ppgn_update} and described as follows:

\begin{itemize}
	\item $\epsilon_1$ term: take a step toward an image that causes the neuron $y_c$ to be the \emph{highest activated} across a layer (Fig.~\ref{fig:ppgn_update}; red arrow)
	\item $\epsilon_2$ term: take a step toward a generic, realistic-looking image (Fig.~\ref{fig:ppgn_update}; blue arrow).
	\item $\epsilon_3$ term: add a small amount of noise to jump around the search space to encourage image diversity (Fig.~\ref{fig:ppgn_update}; green arrow).
\end{itemize}

\noindent\textbf{Maximizing raw activations vs. softmax}
Note that the $\epsilon_1$ term in Eq.~\ref{eqn:update_rule} is not the same as the \emph{gradient of raw activation} term in Eq.~\ref{eqn:am_no_priors}. 
We summarize in Table~\ref{tab:logit_vs_softmax} three variants of computing this $\epsilon_1$ gradient term: (1) derivative of logits; (2) derivative of softmax; and (3) derivative of log of softmax.
Several previous works empirically reported that maximizing raw, pre-softmax activations $a^l_i$ produces better visualizations than directly maximizing the softmax values $s^l_i$ (\tabref{logit_vs_softmax}a vs. b); however, this observation had not been fully justified \cite{simonyan2013deep}.
Nguyen et al. \cite{nguyen2017plug} found the log of softmax gradient term (1) working well empirically; and (2) theoretically justifiable under the probabilistic framework in Section~\ref{sec:framework}.


\definecolor{hilightclr}{rgb}{0.0,0.4,0.6}
{
	\begin{table}[t!h]
	  \center
          \vspace{1em}
		\begin{tabular}{>{\raggedright}m{.5\textwidth} | >{\arraybackslash}m{.4\textwidth} }  
			\hline
			\vspace{0.3em}
			\emph{a. Derivative of raw activations.} Worked well in practice \cite{nguyen2016synthesizing,erhan2009visualizing} but may produce \emph{non-selective} stimuli and is not quite the right term under the probabilistic framework in Sec.~\ref{sec:framework}.
			\vspace{0.3em}
			&
			\begin{center}
				$\begin{aligned}
				{\color{hilightclr} \frac{\partial a^l_i}{\partial x}}
				\end{aligned}$ 
			\end{center}
			\\
			\hline
			\vspace{0.3em}
			\emph{b. Derivative of softmax.} Previously avoided due to poor performance \cite{simonyan2013deep,yosinski2015understanding}, but poor performance may have been due to ill-conditioned optimization rather than the inclusion of logits from other classes. 
			\vspace{0.3em}
			&
			\begin{center}
				$\begin{aligned}
				\frac{\partial s^l_i}{\partial x} = s^l_i\left({\color{hilightclr} \frac{\partial a^l_i}{\partial x}} - \sum_j s^l_j \frac{\partial a^l_j}{\partial x}\right)
				\end{aligned}$ 
			\end{center}
			\\
			\hline
			\emph{c. Derivative of log of softmax.} Correct term under the sampler framework in Sec.~\ref{sec:framework}. Well-behaved under optimization, perhaps due to the $\frac{\partial a^l_i}{\partial x}$ term untouched by the $s^l_i$ multiplier. &
			{
				\begin{center}
					$\begin{aligned}
					\frac{\partial \log s^l_i}{\partial x} & = \frac{\partial \log p(y = y_i | x_t)}{\partial x}  \\
					& = {\color{hilightclr} \frac{\partial a^l_i}{\partial x}} - \frac{\partial}{\partial x} \log \sum_j \exp(a^l_j)
					\end{aligned}$
				\end{center}
			} \\
			\hline
		\end{tabular}
		\vspace*{1em}
		\caption{A comparison of derivatives for use in activation maximization methods. (a) has most commonly been used, (b) has worked in the past but with some difficulty, but (c) is correct under the sampler framework in Sec.~\ref{sec:framework} and \cite{nguyen2017plug}.
		}
		\vspace*{-0.5cm}
		\tablabel{logit_vs_softmax}
	\end{table}
}


We refer readers to \cite{nguyen2017plug} for a more complete derivation and discussion of the above MALA sampler.
Using the update rule in Eq.~\ref{eqn:update_rule}, we will next interpret other AM algorithms in the literature.

\subsection{Interpretation of previous algorithms}

Here, we consider four representative approaches in light of the probabilistic framework:

\begin{enumerate}
	\item AM with no priors \cite{nguyen2015deep,szegedy2013intriguing-properties-of-neural,erhan2009visualizing} (discussed in Sec.~\ref{sec:intro})
	\item AM with a Gaussian prior
	\cite{simonyan2013deep,yosinski2015understanding,wei2015understanding} (discussed in Sec.~\ref{sec:am_priors})
	\item AM with hand-designed priors
	\cite{simonyan2013deep,yosinski2015understanding,nguyen2016multifaceted,wei2015understanding,nguyen2015innovation,mahendran2015visualizing} (discussed in Sec.~\ref{sec:am_priors})
	\item AM in the latent space of generator networks \cite{nguyen2016synthesizing,nguyen2017plug} (discussed in Sec.~\ref{sec:dgnam})
\end{enumerate}

\noindent\textbf{Activation maximization with no priors.}
From \eqnref{update_rule}, if we set $(\epsilon_1, \epsilon_2, \epsilon_3) = (1, 0, 0)$
, we obtain a sampler that follows the neuron gradient directly without contributions from a $p(\x)$ term or the addition of noise. In a high-dimensional space, this results in adversarial or rubbish images \cite{szegedy2013intriguing-properties-of-neural,nguyen2015deep} (as discussed in Sec.~\ref{sec:am_priors}).
 We can also interpret the optimization procedure in \cite{szegedy2013intriguing-properties-of-neural,nguyen2015deep} as a sampler with a non-zero $\epsilon_1$ but with a $p(\x)$ such that $\frac{\partial\log p(\x)}{\partial{\x}} = 0$ i.e. a uniform $p(\x)$ where all images are equally likely.

\noindent\textbf{Activation maximization with a Gaussian prior.} To avoid producing high-frequency images \cite{nguyen2015deep} that are uninterpretable, several works have used $L_2$ decay, which can be thought of as a simple zero-mean Gaussian prior over images \cite{simonyan2013deep,yosinski2015understanding,wei2015understanding}.
From \eqnref{update_rule}, if we define a Gaussian $p(\x)$ centered at the origin (assume the mean image has been subtracted) and set $(\epsilon_1, \epsilon_2, \epsilon_3) = (1, \lambda, 0)$, pulling Gaussian constants into $\lambda$, we obtain the following noiseless update rule:

\beq
\x_{t+1} = (1-\lambda) \x_t + \frac{\partial \log p(y = y_c | \x_t)}{\partial \x_t}
\eqnlabel{update_rule_gaussian}
\eeq

The first term decays the current image slightly toward the origin, as appropriate under a Gaussian image prior, and the second term pulls the image toward higher probability regions for the chosen neuron. Here, the second term is computed as the derivative of the log of a softmax transformation of all activations across a layer (see Table~\ref{tab:logit_vs_softmax}).

\noindent\textbf{Activation maximization with hand-designed priors.} In an effort to outdo the simple Gaussian prior, many works have proposed more creative, hand-designed image priors such as Gaussian blur \cite{yosinski2015understanding}, total variation \cite{mahendran2015visualizing}, jitter, rotate, scale \cite{mordvintsev2015inceptionism}, and data-driven patch priors \cite{wei2015understanding}. 
These priors effectively serve as a simple $p(\x)$ component in Eq.~\ref{eqn:update_rule}.
Note that all previous methods considered under this category are noiseless ($\epsilon_3 = 0$).

\noindent\textbf{Activation maximization in the latent space of generator networks}
To ameliorate the problem of poor mixing in the high-dimensional pixel space \cite{bengio2013better},
several works instead performed optimization in a semantically meaningful, low-dimensional feature space of a generator network \cite{nguyen2016synthesizing,yeh2016semantic,brock2016neural,zhu2016generative,nguyen2017plug}.

That approach can be viewed as re-parameterizing $p(\x)$ as $\int_\h p(\x|\h)p(\h)$, and sampling from the joint probability distribution $p(\h,y)$ instead of $p(\x, y)$, treating $\x$ as a deterministic variable.
That is, the update rule in Eq.~\ref{eqn:update_rule} is now changed into the below:

\beq
\h_{t+1} = \h_t + \epsilon_1\frac{\partial \log p(y = y_c | \h_t)}{\partial \h_t} + \epsilon_2\frac{\partial\log p(\h_t)}{\partial{\h_t}} + N(0, \epsilon_3^2)
\eqnlabel{update_rule_h}
\eeq

In this category, DGN-AM \cite{nguyen2016synthesizing} follows the above rule with ($\epsilon_1$,$\epsilon_2$,$\epsilon_3$) = (1,1,0).\footnote{$\epsilon_3 = 0$ because noise was not used in DGN-AM \cite{nguyen2016synthesizing}.}
Specifically, we hand-designed a $p(\h)$ via clipping and $L_2$ regularization (i.e. a Gaussian prior) to keep the code $\h$ within a ``realistic'' range.
PPGNs follows exactly the update rule in Eq.~\ref{eqn:update_rule_h} with a better $p(\h)$ prior learned via a denoising autoencoder \cite{nguyen2017plug}.
PPGNs produce images with better diversity than DGN-AM \cite{nguyen2017plug}.

\section{Applications of Activation Maximization}
\label{sec:results}	

In this section, we review how one may use activation maximization to understand and explain a pre-trained neural network.
The results below are specifically generated by DGN-AM \cite{nguyen2016synthesizing} and PPGNs \cite{nguyen2017plug} where the authors harnessed a general image generator network to synthesize AM images.\\

\begin{figure}
	\centering
	\includegraphics[width=1.0\columnwidth]{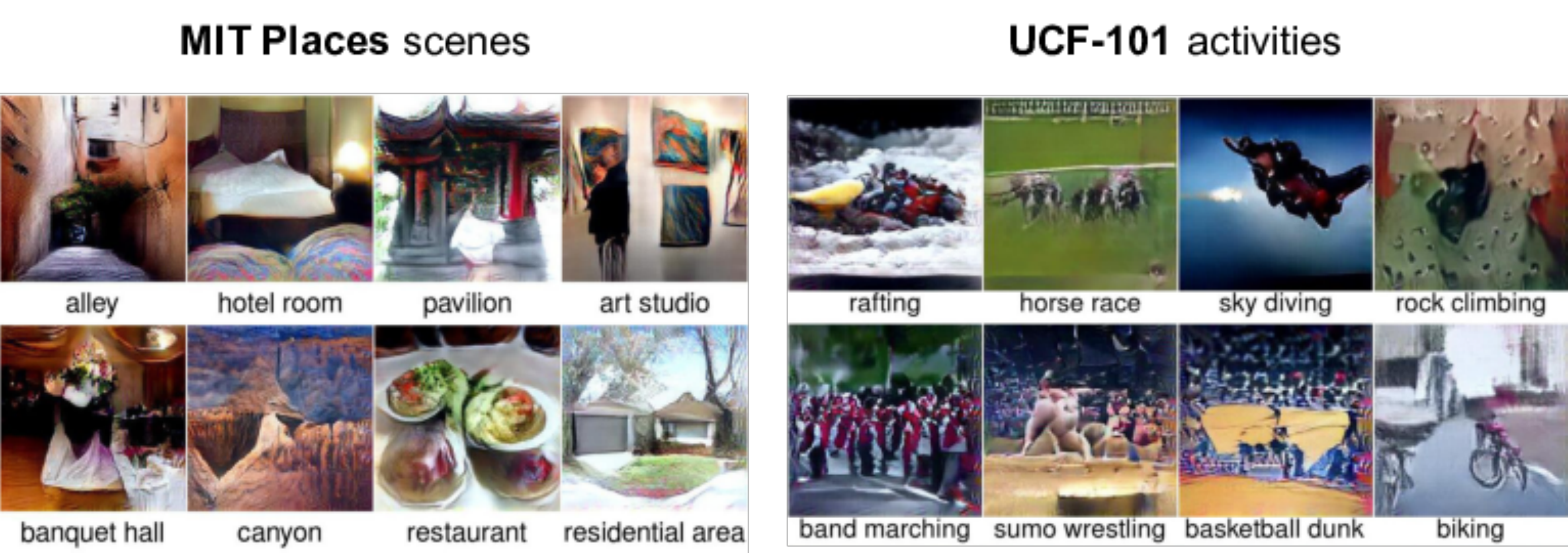}
	\caption{
		Preferred stimuli generated via DGN-AM \cite{nguyen2016synthesizing} for output units of a network trained to classify images on the MIT Places dataset~\cite{zhou2014learning} (left) and a network trained to classify videos from the UCF-101 dataset (right). The results suggested that the learned ImageNet prior generalizes well to synthesizing images for other datasets.
	}
	\label{fig:places205}
\end{figure}

\newpage
\noindent\textbf{Visualize output units for new tasks}
We can harness a general learned ImageNet prior to synthesize images for networks trained on a different dataset e.g. MIT Places dataset \cite{zhou2014object} or UCF-101 activity videos \cite{nguyen2016synthesizing} (Figs.~\ref{fig:teaser} \& \ref{fig:places205}).

\begin{figure}
	\centering
	\includegraphics[width=1.0\textwidth]{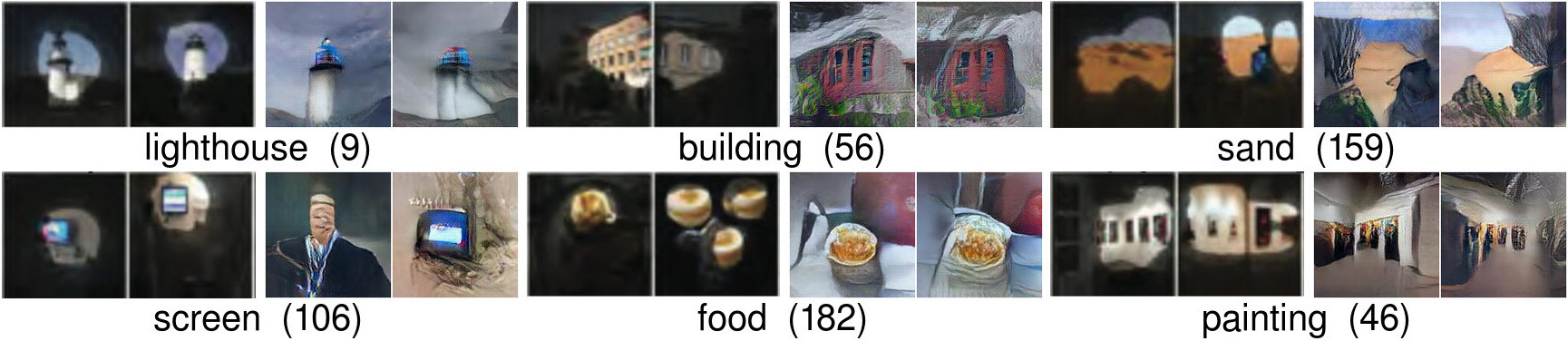}
	\caption{
		AM images for example hidden units at layer \layer{5} of an CaffeNet \cite{krizhevsky2012imagenet} trained to classify images of scenes \cite{zhou2014object}. 
		For each unit: the left two images are masked-out real images, each highlighting a region that highly activates the unit via methods in \cite{zhou2014object}, and humans provide text labels (e.g. ``lighthouse'') describing the common theme in the highlighted regions. 
		The right two images are AM images, which enable the same conclusion regarding what feature a hidden unit has learned. 
		Figure from \cite{nguyen2016synthesizing}.
	}
	\label{fig:deepscene_short}
\end{figure}

\noindent\textbf{Visualize hidden units}
Instead of synthesizing preferred inputs for output neurons (Fig.~\ref{fig:teaser}), one may apply AM to the hidden units.
In a comparison with visualizing \emph{real} image regions that highly activate a unit \cite{zhou2014object}, we found AM images may provide similar but sometimes also complementary evidence suggesting what a unit is for \cite{nguyen2016synthesizing} (Fig.~\ref{fig:deepscene_short}).
For example, via DGN-AM, we found that a unit that detects ``TV screens'' also detects people on TV (Fig.~\ref{fig:deepscene_short}, unit \unit{106}).\\

\begin{figure}[b!h]
	\centering
	\vspace*{-0.5cm}
	\includegraphics[width=0.7\columnwidth]{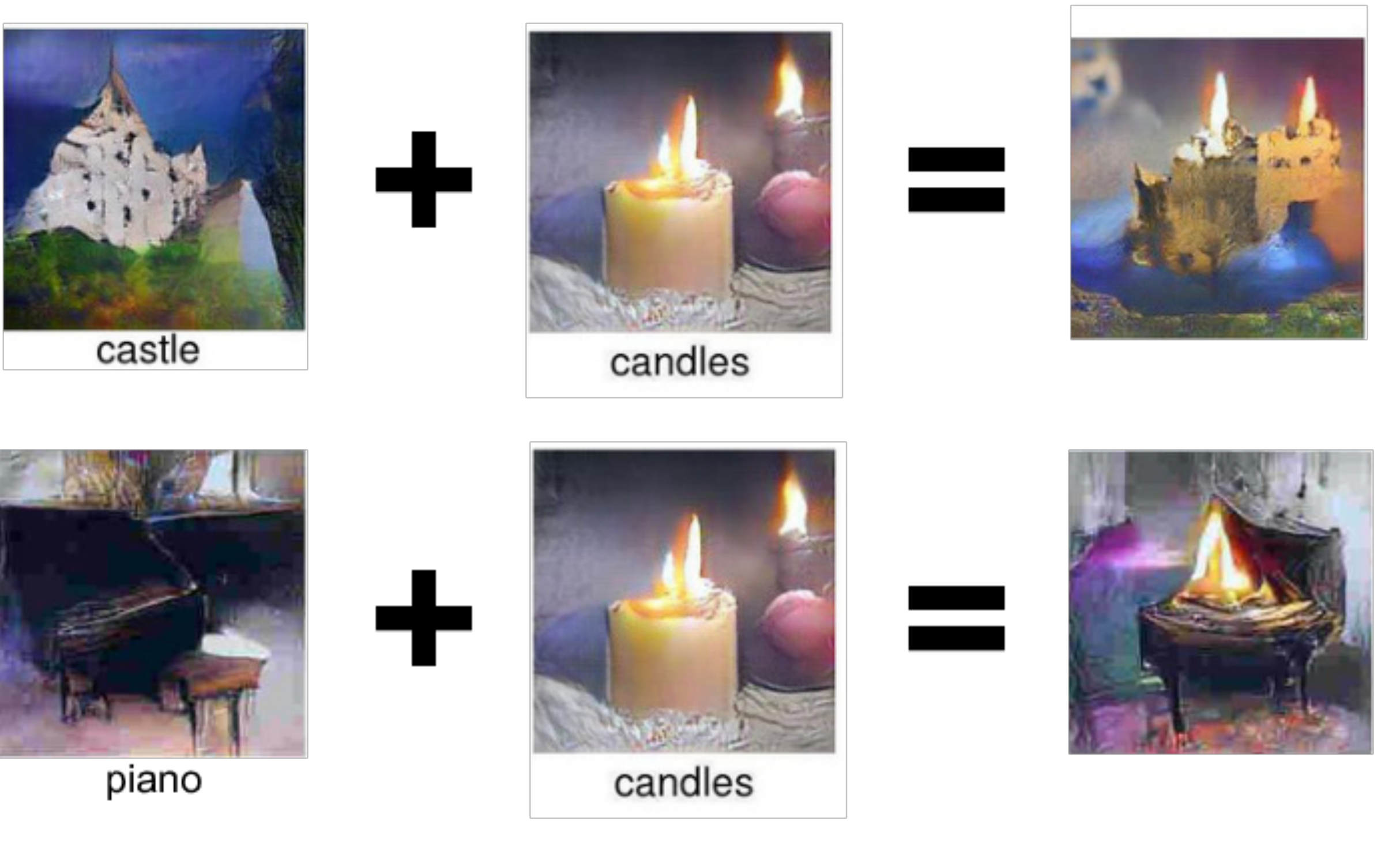}
	\caption{
		Synthesizing images via DGN-AM \cite{nguyen2016synthesizing} to activate both the ``castle'' and ``candles'' units of CaffeNet \cite{krizhevsky2012imagenet} produces an image that resembles a castle on fire (top right).
		Similarly, ``piano'' $\thinplus$ ``candles'' produces a candle on a piano (bottom right).
		Both rightmost images highly activate the ``candles'' output neuron.
	}
	\label{fig:am_math}
	\vspace*{-0.2cm}
\end{figure}

\noindent\textbf{Synthesize preferred images activating multiple neurons}
First, one may synthesize images activating a group of units at the same time to study the interaction between them \cite{nguyen2016synthesizing,olahfeature}.
For example, it might be useful to study how a network distinguishes two related and visually similar concepts such as ``impala'' and ``hartebeest'' animals in ImageNet \cite{deng2009imagenet}.
One way to do this is to synthesize images that maximize the ``impala'' neuron's activation but also \emph{minimize} the ``hartebeest'' neuron's activation.
Second, one may reveal different facets of a neuron \cite{nguyen2016multifaceted} by activating different pairs of units.
That is, activating two units at the same time e.g. (castle + candle); and (piano + candle) would produce two distinct images of candles that activate the same ``candle'' unit \cite{nguyen2016synthesizing} (Fig.~\ref{fig:am_math}).
In addition, this method sometimes also produces interesting, creative art \cite{nguyen2016synthesizing,yahoo2016nsfw}.\\

\noindent\textbf{Watch feature evolution during training}
We can watch how the features evolved during the training of a target classifier network \cite{nguyen2016synthesizing}. 
Example videos of AM visualizations for sample output and hidden neurons during the training of CaffeNet \cite{jia2014caffe} are at: \url{https://www.youtube.com/watch?v=q4yIwiYH6FQ} and \url{https://www.youtube.com/watch?v=G8AtatM1Sts}.
One may find that features at lower layers tend to converge faster vs. those at higher layers.\\

\noindent\textbf{Synthesizing videos}
To gain insights into the inner functions of an activity recognition network \cite{soomro2012ucf101}, one can synthesize a single frame (Fig.~\ref{fig:places205}; right) or an entire \emph{preferred video}.
By synthesizing videos, Nguyen et al. \cite{nguyen2016synthesizing} found that a video recognition network (LRCN \cite{donahue2014long}) classifies videos without paying attention to temporal correlation across video frames.
That is, the AM videos\footnote{https://www.youtube.com/watch?v=IOYnIK6N5Bg} appear to be a set of uncorrelated frames of activity e.g. a basketball game.
Further tests confirmed that the network produces similar top-1 predicted labels regardless of whether the frames of the \emph{original} UCF-101 videos \cite{soomro2012ucf101} are randomly shuffled.\\

\begin{figure*}[!h]
	\centering
	\begin{subfigure}{1.0\linewidth}
		\centering
		\includegraphics[width=1.0\linewidth]{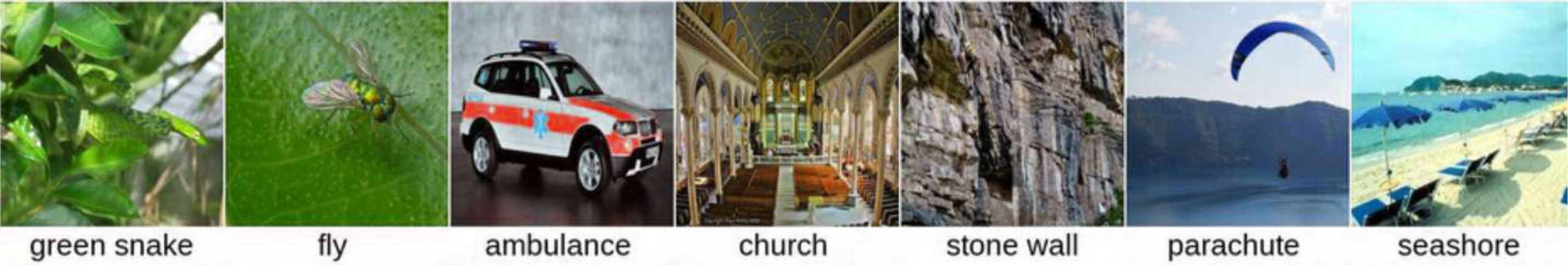}
		\caption{Regular ImageNet training images}
		\vspace{0.2cm}
		
		\label{fig:bug_rgb}
	\end{subfigure}	
	\hspace{1mm}
	\begin{subfigure}{1.0\linewidth}
		\centering
		\includegraphics[width=1.0\linewidth]{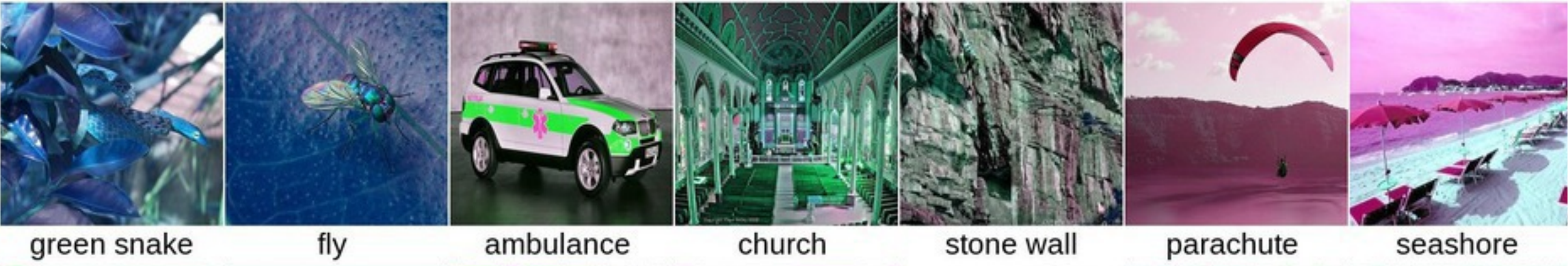}
		\caption{ImageNet training images converted into the BRG color space}
		\vspace{0.2cm}		
		\label{fig:bug_brg}
	\end{subfigure}	
	\hspace{1mm}	
	\begin{subfigure}{1.0\linewidth}
		\centering
		\includegraphics[width=1.0\linewidth]{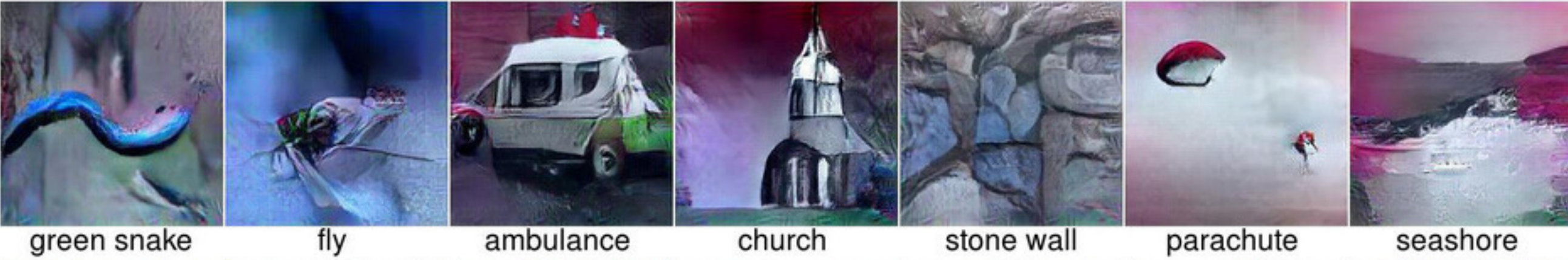}
		\caption{Visualizations of the units that are trained on BRG ImageNet images above (b)}
		\label{fig:bug_am}
	\end{subfigure}
	\caption{
		The original ImageNet training set images are in RGB color space (a).
		We train CaffeNet \cite{krizhevsky2012imagenet} on their BRG versions (b).
		The activation maximization images synthesized by DGN-AM \cite{nguyen2016synthesizing}, faithfully portray the color space of the images, here BRG, where the network was trained on.
	}
	\label{fig:bug}
\end{figure*}

\noindent\textbf{Activation maximization as a debugging tool}
We discuss here a case study where AM can be used as a debugging tool.
Suppose there is a bug in your neural network image classifier implementation that internally and unexpectedly converts all input RGB images (Fig.~\ref{fig:bug_rgb}) into BRG images (Fig.~\ref{fig:bug_brg}) before feeding them to the neural network.
This bug might be hard to notice by only examining accuracy scores or attribution heatmaps \cite{montavon2017methods}.
Instead, AM visualizations could reflect the color space of the images that were fed to the neural network and reveal this bug (Fig.~\ref{fig:bug_am}).

\begin{figure}[t!h]
	\centering
	\includegraphics[width=0.65\columnwidth]{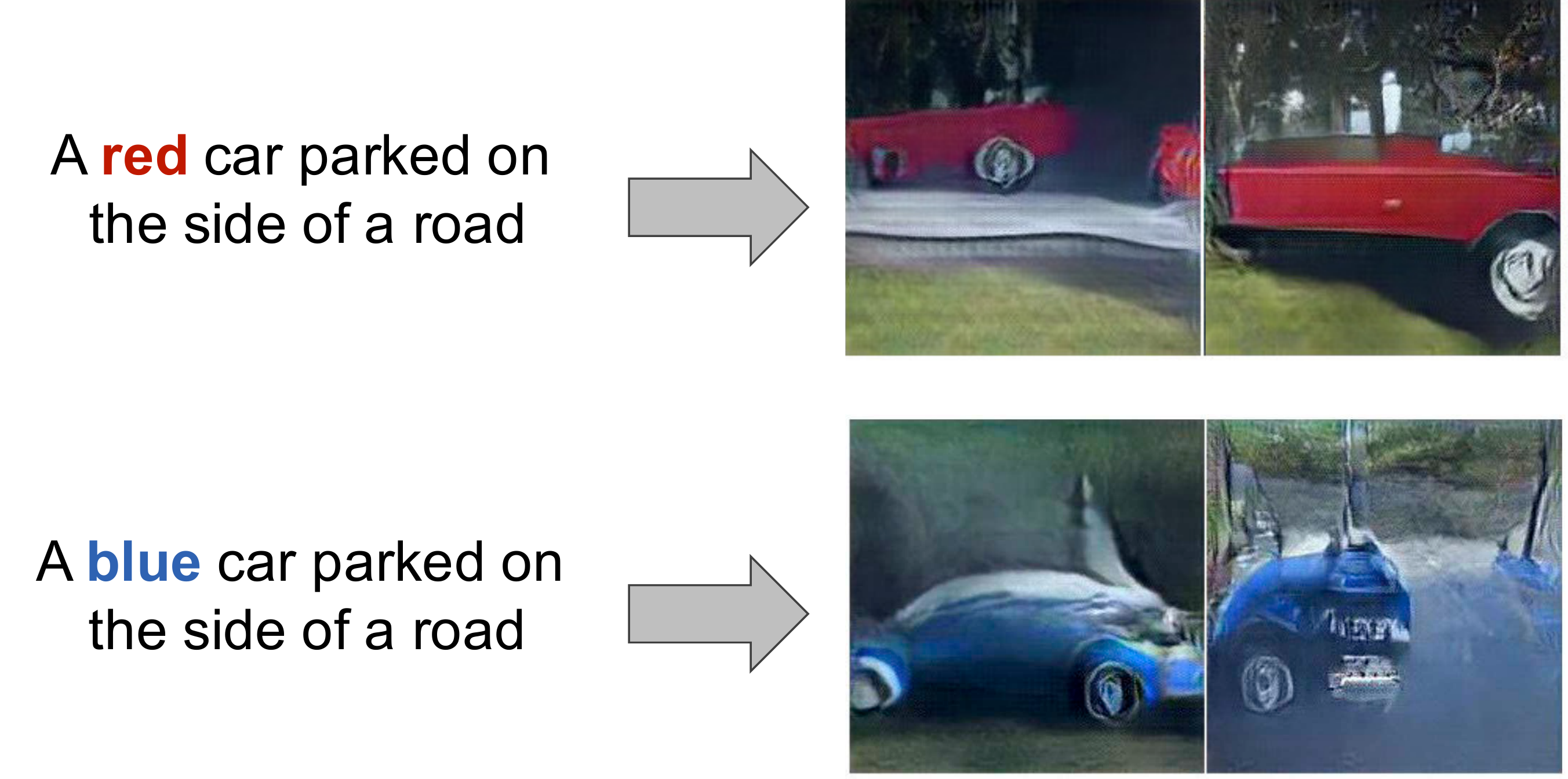}
	\caption{
		We synthesize input images (right) such that a pre-trained image captioning network (LRCN \cite{donahue2014long}) outputs the target caption description (left sentences).
		Each image on the right was produced by starting optimization from a different random initialization.
	}
	\label{fig:blue_car}
	\vspace*{-0.5cm}
\end{figure}

\noindent\textbf{Synthesize preferred images conditioned on a sentence}
Instead of synthesizing images preferred by output units in an image classifier, we can also synthesize images that cause an image \emph{captioning} network to output a desired sentence (examples in Fig.~\ref{fig:blue_car}).

\begin{figure}[t!h]
	\centering
	\includegraphics[width=0.7\columnwidth]{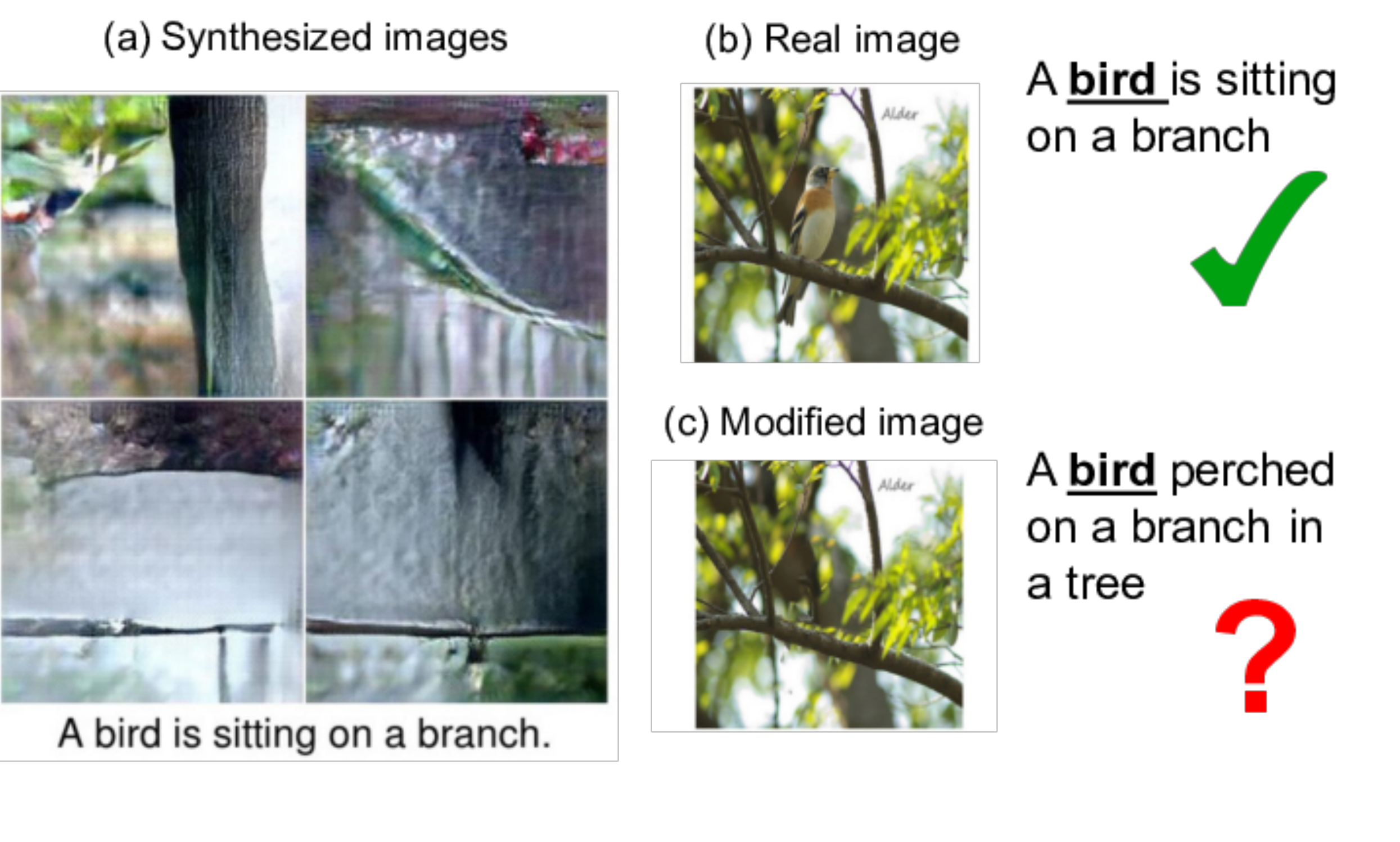}
	\caption{
		While synthesizing images to cause an image captioning model \cite{donahue2014long} to output  \emph{``A bird is sitting on a branch''} via DGN-AM method \cite{nguyen2016synthesizing}, we only obtained images of branches or trees that surprisingly has no birds at all (a).
		Further tests on real MS COCO images revealed that the model \cite{donahue2014long} outputs correct captions for a test image that has a bird (b), but still insists on the existence of the bird, even when it is manually removed via Adobe Photoshop (c).
		This suggests the image captioner learned a strong correlation between birds and tree branches---a bias that might exist in the language or image model.
	}
	\label{fig:bird}
\end{figure}

This reverse-engineering process may uncover interesting insights into the system's behaviors. 
For example, we discovered an interesting failure of a state-of-the-art image captioner \cite{donahue2014long} when it declares birds even when there is no bird in an image (Fig.~\ref{fig:bird}).

\begin{figure*}[!h]
	\centering
	\includegraphics[width=0.83\columnwidth]{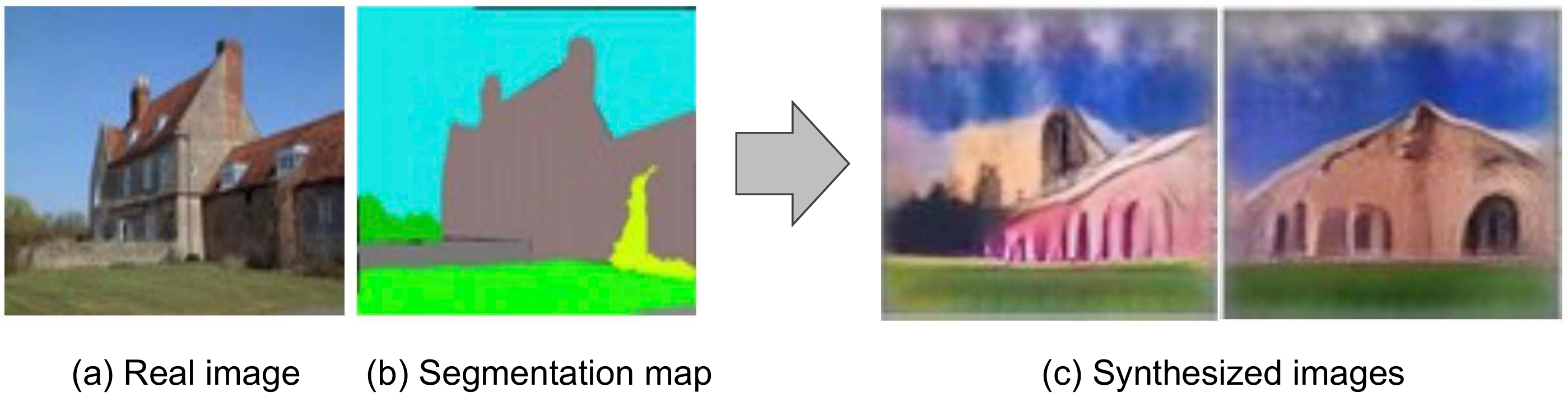}
	\caption{
		A segmentation network from \cite{zhou2017scene} is capable of producing a semantic segmentation map (b) given an input real image (a).
		The authors extend the DGN-AM method \cite{nguyen2016synthesizing} to synthesize images (c) to match the target segmentation map (b), which specifies a scene with a building on green grass and under a blue sky background.
		Figure modified from \cite{zhou2017scene}.
	}
	\vspace*{-0.5cm}
	\label{fig:segmentation}
\end{figure*}

\newpage
\noindent\textbf{Synthesize preferred images conditioned on a semantic segmentation map}
We can extend AM methods to synthesize images with more fine-grained controls of where objects are placed by matching a semantic map output of a segmentation network (Fig.~\ref{fig:segmentation}) or a target spatial feature map of a convolutional layer.

\noindent\textbf{Synthesize preferred stimuli for real, biological brains}
While this survey aims at visualizing artificial networks, it is also possible to harness our AM techniques to study biological brains.
Two teams of Neuroscientists \cite{malakhova-2018-JOpt-visualization-of-information-encoded,ponce2019evolving} have recently been able to reconstruct stimuli for neurons in alive macaques' brains using either the ImageNet PPGN (as discussed in Sec.~\ref{sec:probabilistic}) \cite{malakhova-2018-JOpt-visualization-of-information-encoded} or the DGN-AM (as discussed in Sec.~\ref{sec:dgnam}) \cite{ponce2019evolving}.
The synthesized images surprisingly resemble monkeys and human nurses that the subject macaque meets frequently \cite{ponce2019evolving} or show eyes in neurons previously shown to be tuned for detecting faces \cite{malakhova-2018-JOpt-visualization-of-information-encoded}.
Similar AM frameworks have also been interestingly applied to reconstruct stimuli from EEG or MRI signals of human brains \cite{shen2019deep,palazzo2018decoding}.

\section{Discussion and Conclusion}

While activation maximization has proven a useful tool for understanding neural networks, there are still open challenges and opportunities such as:

\begin{itemize}
	\item One might wish to harness AM to compare and contrast the features learned by different models.
	That would require a robust, principled AM approach that produces faithful and interpretable visualizations of the learned features for networks trained on different datasets or of different architectures.
	This is challenging due to two problems: (1) the image prior may not be general enough and may have a bias toward a target network or one dataset over the others; (2) AM optimization on different network architectures, especially of different depths, often requires different hyper-parameter settings to obtain the best performance.
	\item It is important for the community to propose rigorous approaches for evaluating AM methods.
	A powerful image prior may incur a higher risk of producing misleading visualizations---it is unclear whether a synthesized visual feature comes from the image prior or the target network being studied or both.
	Note that we have investigated that and surprisingly found the DGN-AM prior to be able to generate a wide diversity of images including the non-realistic ones (e.g. blurry, cut-up, and BRG images \cite{nguyen2016synthesizing}).
	\item Concepts in modern deep networks can be highly distributed \cite{fong2018net2vec,bau2017network,szegedy2013intriguing-properties-of-neural}; therefore, it might be promising to apply AM to study networks at a different, larger scale than individual neurons, e.g. looking at groups of neurons \cite{olah2018building}.	
	\item It might be a fruitful direction to combine AM with other tools such as attribution heatmapping \cite{olah2018building} or integrate AM into the testbeds for AI applications \cite{pei2017deepxplore} as we move towards safe, transparent, and fair AI.
	\item One may also perform AM in the parameter space of a 3D renderer (e.g. modifying the lighting, object geometry or appearances in a 3D scene) that renders a 2D image that strongly activates a unit \cite{alcorn2019strike}.
	AM in a 3D space allows us to synthesize stimuli by varying a controlled factor (e.g. lighting) and thus might offer deeper insights into a model's inner-workings.
\end{itemize}

Activation maximization techniques enable us to shine light into the black-box neural networks. 
As this survey shows, improving activation maximization techniques improves our ability to understand deep neural networks. 
We are excited for what the future holds regarding improved techniques that make neural networks more interpretable and less opaque so we can better understand how deep neural networks do the amazing things that they do. 



\subsubsection*{Acknowledgments}
Anh Nguyen is supported by Amazon Research Credits, Auburn University, and donations from Adobe Systems Inc and Nvidia.


\bibliographystyle{splncs03}
\bibliography{references}

\begin{thebibliography}{10}
\providecommand{\url}[1]{\texttt{#1}}
\providecommand{\urlprefix}{URL }

\bibitem{akhtar2018threat}
Akhtar, N., Mian, A.: Threat of adversarial attacks on deep learning in
  computer vision: A survey. IEEE Access  6,  14410--14430 (2018)

\bibitem{alcorn2019strike}
Alcorn, M.A., Li, Q., Gong, Z., Wang, C., Mai, L., Ku, W.S., Nguyen, A.: Strike
  (with) a pose: Neural networks are easily fooled by strange poses of familiar
  objects. In: Proceedings of the IEEE Conference on Computer Vision and
  Pattern Recognition. vol.~1, p.~4. IEEE (2019)

\bibitem{baer2007neuroscience}
Baer, M., Connors, B.W., Paradiso, M.A.: Neuroscience: Exploring the brain
  (2007)

\bibitem{bau2017network}
Bau, D., Zhou, B., Khosla, A., Oliva, A., Torralba, A.: Network dissection:
  Quantifying interpretability of deep visual representations. In: Computer
  Vision and Pattern Recognition (CVPR), 2017 IEEE Conference on. pp.
  3319--3327. IEEE (2017)

\bibitem{bengio2013better}
Bengio, Y., Mesnil, G., Dauphin, Y., Rifai, S.: Better mixing via deep
  representations. In: International Conference on Machine Learning. pp.
  552--560 (2013)

\bibitem{brock2016neural}
Brock, A., Lim, T., Ritchie, J.M., Weston, N.: Neural photo editing with
  introspective adversarial networks. arXiv preprint arXiv:1609.07093  (2016)

\bibitem{deng2009imagenet}
Deng, J., et~al.: Imagenet: A large-scale hierarchical image database. In: CVPR
  (2009)

\bibitem{donahue2014long}
Donahue, J., Hendricks, L.A., Guadarrama, S., Rohrbach, M., et~al.: Long-term
  recurrent convolutional networks for visual recognition and description. In:
  Computer Vision and Pattern Recognition (2015)

\bibitem{dosovitskiy2016generating}
Dosovitskiy, A., Brox, T.: Generating images with perceptual similarity metrics
  based on deep networks. In: NIPS (2016)

\bibitem{erhan2009visualizing}
Erhan, D., Bengio, Y., Courville, A., Vincent, P.: Visualizing higher-layer
  features of a deep network. Dept. IRO, Universit{\'e} de Montr{\'e}al, Tech.
  Rep  4323 (2009)

\bibitem{fong2018net2vec}
Fong, R., Vedaldi, A.: Net2vec: Quantifying and explaining how concepts are
  encoded by filters in deep neural networks. arXiv preprint arXiv:1801.03454
  (2018)

\bibitem{yahoo2016nsfw}
Goh, G.: Image synthesis from {Y}ahoo {O}pen {NSFW}.
  \url{https://opennsfw.gitlab.io} (2016)

\bibitem{goodfellow2014generative}
Goodfellow, I., Pouget-Abadie, J., Mirza, M., Xu, B., Warde-Farley, D., Ozair,
  S., Courville, A., Bengio, Y.: Generative adversarial nets. In: NIPS (2014)

\bibitem{hubel1959receptive}
Hubel, D.H., Wiesel, T.N.: Receptive fields of single neurones in the cat's
  striate cortex. The Journal of physiology  148(3),  574--591 (1959)

\bibitem{jia2014caffe}
Jia, Y., Shelhamer, E., Donahue, J., Karayev, S., Long, J., Girshick, R.,
  Guadarrama, S., Darrell, T.: Caffe: Convolutional architecture for fast
  feature embedding. arXiv preprint arXiv:1408.5093  (2014)

\bibitem{kabilan2018vectordefense}
Kabilan, V.M., Morris, B., Nguyen, A.: Vectordefense: Vectorization as a
  defense to adversarial examples. arXiv preprint arXiv:1804.08529  (2018)

\bibitem{kandel2000principles}
Kandel, E.R., Schwartz, J.H., Jessell, T.M., Siegelbaum, S.A., Hudspeth, A.J.,
  et~al.: Principles of neural science, vol.~4. McGraw-hill New York (2000)

\bibitem{krizhevsky2012imagenet}
Krizhevsky, A., Sutskever, I., Hinton, G.E.: Imagenet classification with deep
  convolutional neural networks. In: Advances in neural information processing
  systems. pp. 1097--1105 (2012)

\bibitem{le2013building}
Le, Q.V.: Building high-level features using large scale unsupervised learning.
  In: Acoustics, Speech and Signal Processing (ICASSP), 2013 IEEE International
  Conference on. pp. 8595--8598. IEEE (2013)

\bibitem{li2015convergent}
Li, Y., Yosinski, J., Clune, J., Lipson, H., Hopcroft, J.: Convergent learning:
  Do different neural networks learn the same representations? In: Feature
  Extraction: Modern Questions and Challenges. pp. 196--212 (2015)

\bibitem{mahendran2015visualizing}
Mahendran, A., Vedaldi, A.: Visualizing deep convolutional neural networks
  using natural pre-images. In: Computer Vision and Pattern Recognition (CVPR)
  (2016)

\bibitem{malakhova-2018-JOpt-visualization-of-information-encoded}
Malakhova, K.: Visualization of information encoded by neurons in the
  higher-level areas of the visual system. J. Opt. Technol.  85(8),  494--498
  (Aug 2018)

\bibitem{montavon2017methods}
Montavon, G., Samek, W., M{\"u}ller, K.R.: Methods for interpreting and
  understanding deep neural networks. Digital Signal Processing  (2017)

\bibitem{mordvintsev2015inceptionism}
Mordvintsev, A., Olah, C., Tyka, M.: Inceptionism: Going deeper into neural
  networks. Google Research Blog. Retrieved June  20 (2015)

\bibitem{nguyen2017ai}
Nguyen, A., of~Wyoming. Computer Science~Department, U.: AI Neuroscience:
  Visualizing and Understanding Deep Neural Networks. University of Wyoming
  (2017), \url{https://books.google.com/books?id=QCexswEACAAJ}

\bibitem{nguyen2017plug}
Nguyen, A., Clune, J., Bengio, Y., Dosovitskiy, A., Yosinski, J.: Plug \& play
  generative networks: Conditional iterative generation of images in latent
  space. In: Computer Vision and Pattern Recognition (CVPR), 2017 IEEE
  Conference on. pp. 3510--3520. IEEE (2017)

\bibitem{nguyen2016synthesizing}
Nguyen, A., Dosovitskiy, A., Yosinski, J., Brox, T., Clune, J.: Synthesizing
  the preferred inputs for neurons in neural networks via deep generator
  networks. In: Advances in Neural Information Processing Systems. pp.
  3387--3395 (2016)

\bibitem{nguyen2015deep}
Nguyen, A., Yosinski, J., Clune, J.: Deep neural networks are easily fooled:
  High confidence predictions for unrecognizable images. In: Computer Vision
  and Pattern Recognition (CVPR) (2015)

\bibitem{nguyen2016multifaceted}
Nguyen, A., Yosinski, J., Clune, J.: Multifaceted feature visualization:
  Uncovering the different types of features learned by each neuron in deep
  neural networks. In: Visualization for Deep Learning Workshop, ICML
  conference (2016)

\bibitem{nguyen2016understanding}
Nguyen, A., Yosinski, J., Clune, J.: Understanding innovation engines:
  Automated creativity and improved stochastic optimization via deep learning.
  Evolutionary Computation  24(3),  545--572 (2016)

\bibitem{nguyen2015innovation}
Nguyen, A.M., Yosinski, J., Clune, J.: Innovation engines: Automated creativity
  and improved stochastic optimization via deep learning. In: Proceedings of
  the 2015 Annual Conference on Genetic and Evolutionary Computation. pp.
  959--966. ACM (2015)

\bibitem{olahfeature}
Olah, C., Mordvintsev, A., Schubert, L.: Feature visualization. distill, 2017.
  doi: 10.23915/distill. 00007

\bibitem{olah2018building}
Olah, C., Satyanarayan, A., Johnson, I., Carter, S., Schubert, L., Ye, K.,
  Mordvintsev, A.: The building blocks of interpretability. Distill  3(3),  e10
  (2018)

\bibitem{palazzo2018decoding}
Palazzo, S., Spampinato, C., Kavasidis, I., Giordano, D., Shah, M.: Decoding
  brain representations by multimodal learning of neural activity and visual
  features. arXiv preprint arXiv:1810.10974  (2018)

\bibitem{pei2017deepxplore}
Pei, K., Cao, Y., Yang, J., Jana, S.: Deepxplore: Automated whitebox testing of
  deep learning systems. In: Proceedings of the 26th Symposium on Operating
  Systems Principles. pp. 1--18. ACM (2017)

\bibitem{ponce2019evolving}
Ponce, C.R., Xiao, W., Schade, P., Hartmann, T.S., Kreiman, G., Livingstone,
  M.S.: Evolving super stimuli for real neurons using deep generative networks.
  bioRxiv p. 516484 (2019)

\bibitem{quiroga2005invariant}
Quiroga, R.Q., Reddy, L., Kreiman, G., Koch, C., Fried, I.: Invariant visual
  representation by single neurons in the human brain. Nature  435(7045),
  1102--1107 (2005)

\bibitem{roberts1998optimal}
Roberts, G.O., Rosenthal, J.S.: Optimal scaling of discrete approximations to
  langevin diffusions. Journal of the Royal Statistical Society: Series B
  (Statistical Methodology)  60(1),  255--268 (1998)

\bibitem{rumelhart1986learning}
Rumelhart, D.E., Hinton, G.E., Williams, R.J.: Learning representations by
  back-propagating errors. nature  323(6088),  533 (1986)

\bibitem{russakovsky2014imagenet}
Russakovsky, O., et~al.: Imagenet large scale visual recognition challenge.
  IJCV  115(3),  211--252 (2015)

\bibitem{shen2019deep}
Shen, G., Horikawa, T., Majima, K., Kamitani, Y.: Deep image reconstruction
  from human brain activity. PLoS computational biology  15(1),  e1006633
  (2019)

\bibitem{simonyan2013deep}
Simonyan, K., Vedaldi, A., Zisserman, A.: Deep inside convolutional networks:
  Visualising image classification models and saliency maps. ICLR workshop
  (2014)

\bibitem{soomro2012ucf101}
Soomro, K., Zamir, A.R., Shah, M.: Ucf101: A dataset of 101 human actions
  classes from videos in the wild. arXiv preprint arXiv:1212.0402  (2012)

\bibitem{szegedy2013intriguing-properties-of-neural}
Szegedy, C., Zaremba, W., Sutskever, I., Bruna, J., Erhan, D., Goodfellow,
  I.J., Fergus, R.: Intriguing properties of neural networks. CoRR
  abs/1312.6199 (2013)

\bibitem{tyka2016bilateral}
Tyka, M.: Class visualization with bilateral filters.
  \url{https://mtyka.github.io/deepdream/2016/02/05/bilateral-class-vis.html},
  (Accessed on 06/26/2018)

\bibitem{wei2015understanding}
Wei, D., Zhou, B., Torrabla, A., Freeman, W.: Understanding intra-class
  knowledge inside cnn. arXiv preprint arXiv:1507.02379  (2015)

\bibitem{yeh2016semantic}
Yeh, R., Chen, C., Lim, T.Y., Hasegawa-Johnson, M., Do, M.N.: Semantic image
  inpainting with perceptual and contextual losses. arxiv preprint. arXiv
  preprint arXiv:1607.07539  2 (2016)

\bibitem{yosinski2015understanding}
Yosinski, J., Clune, J., Nguyen, A., Fuchs, T., Lipson, H.: Understanding
  neural networks through deep visualization. In: Deep Learning Workshop, ICML
  conference (2015)

\bibitem{zeiler2014visualizing}
Zeiler, M.D., Fergus, R.: Visualizing and understanding convolutional networks.
  In: Computer Vision--ECCV 2014, pp. 818--833. Springer (2014)

\bibitem{zhou2014object}
Zhou, B., Khosla, A., Lapedriza, A., Oliva, A., Torralba, A.: Object detectors
  emerge in deep scene cnns. In: International Conference on Learning
  Representations (ICLR) (2015)

\bibitem{zhou2014learning}
Zhou, B., Lapedriza, A., Xiao, J., Torralba, A., Oliva, A.: Learning deep
  features for scene recognition using places database. In: Advances in neural
  information processing systems (2014)

\bibitem{zhou2017scene}
Zhou, B., Zhao, H., Puig, X., Fidler, S., Barriuso, A., Torralba, A.: Scene
  parsing through ade20k dataset. In: Proceedings of the IEEE Conference on
  Computer Vision and Pattern Recognition. vol.~1, p.~4. IEEE (2017)

\bibitem{zhu2016generative}
Zhu, J.Y., Kr{\"a}henb{\"u}hl, P., Shechtman, E., Efros, A.A.: Generative
  visual manipulation on the natural image manifold. In: European Conference on
  Computer Vision. pp. 597--613. Springer (2016)

\bibitem{oygard2015visualizing}
Øygard, A.M.: Visualizing {G}oog{L}e{N}et classes | audun m {\o}ygard.
  \url{https://www.auduno.com/2015/07/29/visualizing-googlenet-classes/},
  (Accessed on 06/26/2018)

\end{thebibliography}


\end{document}